\documentclass{article}

\PassOptionsToPackage{numbers,sort&compress}{natbib}
\usepackage[preprint]{neurips_2026}

\usepackage[utf8]{inputenc} % allow utf-8 input
\usepackage[T1]{fontenc}    % use 8-bit T1 fonts
\usepackage{hyperref}       % hyperlinks
\usepackage{url}            % simple URL typesetting
\usepackage{booktabs}       % professional-quality tables
\usepackage{amsfonts}       % blackboard math symbols
\usepackage{nicefrac}       % compact symbols for 1/2, etc.
\usepackage{microtype}      % microtypography
\usepackage{xcolor}         % colors
\usepackage{multirow}
\usepackage{pifont}
\usepackage[table,xcdraw]{xcolor}
\usepackage{graphicx}
\usepackage{amsmath}
\usepackage{enumitem}
\usepackage[normalem]{ulem}
\usepackage[ruled]{algorithm2e}
\usepackage[table]{xcolor}
\usepackage{multirow}
\usepackage{amssymb}

\title{What Limits Vision-and-Language Navigation ?}

\author{
    \textbf{Yunheng Wang}\textsuperscript{1},~
    \textbf{Yuetong Fang}\textsuperscript{1},~
    \textbf{Taowen Wang}\textsuperscript{1},~
    \textbf{Lusong Li}\textsuperscript{2},~
    \textbf{Kun Liu}\textsuperscript{2},~
    \\
    \textbf{Junzhe Xu}\textsuperscript{1,2},~
    \textbf{Zizhao Yuan}\textsuperscript{1},~
    \textbf{Yixiao Feng}\textsuperscript{1},~
    \textbf{Jiaxi Zhang}\textsuperscript{1},~
    \\
    \textbf{Wei Lu}\textsuperscript{1},~
    \textbf{Zecui Zeng}\textsuperscript{2,\textdagger},~
    \textbf{Renjing Xu}\textsuperscript{1,\textdagger}
    \\
    \\
    \textsuperscript{1}{\small HKUST(GZ)}~
    \textsuperscript{2}{\small JD Explore Academy}
}

\begin{document}

\maketitle

% 预印文本时候开启
\vspace{-0.2em}

\begin{abstract} 
Vision-and-Language Navigation (VLN) is a cornerstone of embodied intelligence. However, current agents often suffer from significant performance degradation when transitioning from simulation to real-world deployment, primarily due to perceptual instability (e.g., lighting variations and motion blur) and under-specified instructions. While existing methods attempt to bridge this gap by scaling up model size and training data, we argue that the bottleneck lies in the lack of robust spatial grounding and cross-domain priors.
In this paper, we propose StereoNav, a robust Vision-Language-Action framework designed to enhance real-world navigation consistency. To address the inherent gap between synthetic training and physical execution, we introduce Target-Location Priors as a persistent bridge. These priors provide stable visual guidance that remains invariant across domains, effectively grounding the agent even when instructions are vague. Furthermore, to mitigate visual disturbances like motion blur and illumination shifts, StereoNav leverages stereo vision to construct a unified representation of semantics and geometry, enabling precise action prediction through enhanced depth awareness.
Extensive experiments on R2R-CE and RxR-CE demonstrate that StereoNav achieves state-of-the-art egocentric RGB performance, with SR and SPL scores of 81.1\% and 68.3\%, and 67.5\% and 52.0\%, respectively, while using significantly fewer parameters and less training data than prior scaling-based approaches. More importantly, real-world robotic deployments confirm that StereoNav substantially improves navigation reliability in complex, unstructured environments. Project page: \href{https://yunheng-wang.github.io/stereonav-public.github.io/}{\textcolor{blue}{Link}}.

\end{abstract}

\section{Introduction}

Vision-and-Language Navigation (VLN)~\cite{vln} is a cornerstone of embodied intelligence, requiring agents to ground natural language instructions into sequential egocentric observations~\cite{vlntatasiteofm}. Ideally, a robust VLN system should serve as a dependable primitive for advanced downstream tasks, such as mobile manipulation~\cite{HomeRobot} and loco-manipulation~\cite{NaVILA}, where success critically depends on reaching a target with precise spatial alignment. However, despite significant progress in simulation, current VLN agents often exhibit brittle behavior and execution inconsistency when deployed in real-world environments. They struggle to maintain performance amidst the complexities of physical reality, such as varying lighting, camera shake, and unstructured layouts.

% Vision-and-Language Navigation (VLN)~\cite{vln} aims to enable embodied agents to navigate in complex environments by following human natural-language instructions grounded in sequential egocentric observations, and has long been regarded as a fundamental capability for embodied intelligence~\cite{vlntatasiteofm}. In principle, such a capability should serve as a building block for more advanced embodied tasks, such as loco-manipulation~\cite{NaVILA,HLAMCPACICPAL} and mobile manipulation~\cite{OK-Robot,HomeRobot}, where task success critically depends on reaching the correct place with appropriate spatial alignment. However, despite this original motivation, VLN has rarely evolved into a dependable primitive for downstream embodied applications. Current VLN agents still suffer from degraded navigation performance and reduced execution consistency. Their behavior remains brittle even in simulation, and these limitations become more pronounced when deployed in real physical environments, making them difficult to serve as reliable primitives for practical embodied systems.

\begin{figure*}[!t]
    \centering
    \vspace{-1.0em}
    \includegraphics[width=\linewidth]{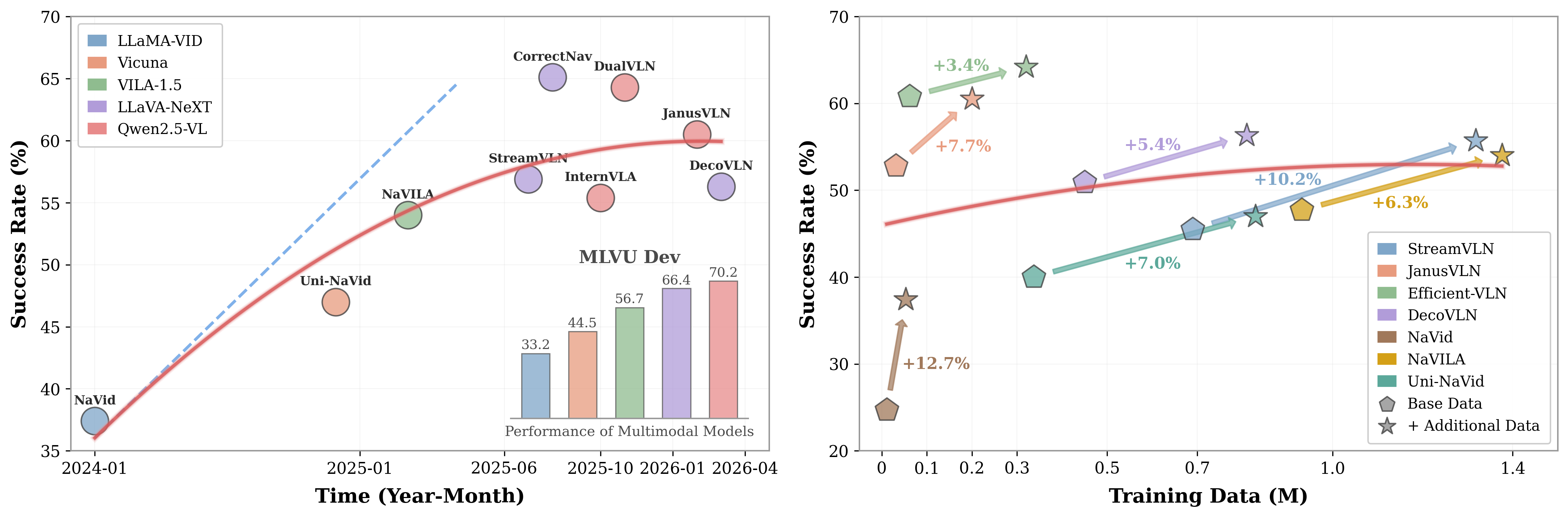}
    \caption{\textbf{Performance under different backbones and training data scales.} (a) \textbf{Left:} The red trend line shows that success rate increases over time with diminishing returns. The inset reports the MLVU-Dev~\cite{Mlvu} scores of the adopted backbone VLMs, indicating that stronger backbones improve early performance, but their gains gradually saturate. (b) \textbf{Right:} Arrows indicate the performance gains from increased training data for each method, while the red trend line suggests that training data scale has only a weak overall correlation with navigation performance across methods.}
    \vspace{-1.0em}
    \label{fig:intro}
\end{figure*}

A prevailing consensus in the community attributes these failures to insufficient cross-modal understanding, leading to a paradigm of "scaling": upgrading to larger and stronger Vision-Language Model (VLM) backbones~\cite{NaVid,NaVILA} and expanding training corpora with auxiliary data~\cite{StreamVLN}. As illustrated in Fig.~\ref{fig:intro}, while these strategies have pushed the state-of-the-art, they are reaching a point of diminishing returns. Models built on comparable backbones show saturated performance, and the correlation between training data scale and success remains surprisingly weak. This suggests that the primary bottleneck may not be "understanding" alone, but rather the lack of robust structural priors that can bridge the gap between abstract instructions and noisy physical perceptions.

% A prevailing consensus in VLN is that these issues primarily stem from insufficient understanding ability of the agent. Under this consensus, prior efforts have largely followed two scaling paths: upgrading the backbone to stronger VLMs~\cite{NaVid,Uni-NaVid,NaVILA,StreamVLN,CorrectNav,DualVLN,JanusVLN,DecoVLN,internvla-n1} and enlarging the training corpus with more navigation and auxiliary data~\cite{StreamVLN, JanusVLN,Efficient-VLN,DecoVLN,NaVid,NaVILA,Uni-NaVid}. As illustrated in Fig. ~\ref{fig:intro}, although both strategies lead to performance improvements, their gains are increasingly limited. Stronger backbones raise the overall trend, yet methods built on comparable VLMs seldom show qualitative gaps, indicating diminishing returns from model scaling alone. Similarly, scaling up training data usually yields only modest improvements, and the overall relationship between data volume and navigation performance remains far from clear. In fact, some methods trained with less data still achieve superior results. This leads to a fundamental question: is insufficient understanding still the primary bottleneck in VLN, or are we overlooking more essential limiting factors?

In this paper, we argue that real-world performance degradation stems from two neglected factors: environmental perception instability and instructional under-specification. In physical settings, egocentric observations are frequently corrupted by motion blur and illumination shifts, which undermine action prediction in models trained on pristine synthetic data. Furthermore, natural language instructions are inherently sparse, providing insufficient guidance for long-range navigation. While humans often navigate using a mental coarse map alongside visual cues, current agents are forced to infer implicit spatial goals solely from ambiguous text, a task that remains difficult to scale through data alone.

% To answer this question, we revisit VLN through the lens of uncertainty. Our analysis suggests that the limitations of current agents do not stem solely from insufficient understanding, but also from two overlooked factors that become especially critical in real-world embodied navigation: visual uncertainty and instruction ambiguity. Egocentric observations are easily disturbed in realistic deployment, and severe perturbations can sharply undermine stable perception and action prediction. At the same time, navigation instructions are often under-specified, inducing ambiguity in route following and goal stopping. Together, these factors reveal a mismatch between current VLN formulations and real-world embodied navigation, indicating that stronger vision-language modeling alone is unlikely to fully resolve the problem.

We propose StereoNav, a stereo Vision-Language-Action framework designed for robust embodied navigation. Addressing the challenge of under-specified instructions, we introduce Target-Point Priors that are derived from the coarse or precise knowledge typically available in practical scenarios. By rendering these priors as persistent visual cues, StereoNav equips the agent with global information guidance, ensuring the goal remains spatially anchored despite incomplete linguistic descriptions. Seeking to mitigate perceptual instability (e.g., lighting variations and motion blur), our framework moves beyond conventional monocular perception by incorporating stereo observations. This approach enables unified semantic, structural, and geometric modeling, allowing the agent to maintain stable perception and reliable decision-making through joint action and depth prediction. Our contributions are threefold:

\begin{itemize}[leftmargin=1.5em, itemsep=2pt, topsep=2pt]
\item We identify perceptual instability (e.g., lighting and motion blur) and instruction under-specification as key factors that hinder the Sim-to-Real transfer of VLN agents.

%, offering a fresh perspective on the execution drift seen in physical environments.

%We revisit VLN from the perspective of practical embodied deployment and identify visual uncertainty and instruction ambiguity as two overlooked factors that fundamentally limit current agents, providing a new explanation for their degraded navigation performance and reduced execution consistency in real-world environments.
    
\item We propose StereoNav, which integrates target-point-prior-guided navigation to provide persistent global guidance and stereo-based geometric modeling to ensure robust environmental perception.

%We propose StereoNav, a stereo Vision-Language-Action framework that addresses these limitations through target-point-prior-guided navigation and stereo-based unified understanding with joint action and depth prediction for more effective embodied decision making.
    
\item We demonstrate that StereoNav achieves state-of-the-art performance on competitive benchmarks while exhibiting significantly higher deployment reliability and execution consistency in real-world robotic tasks compared to monocular alternatives.

%We show that StereoNav achieves state-of-the-art results on both R2R/RxR-CE with substantially fewer parameters and less training data than prior methods, while also exhibiting stronger robustness and higher reliability in real-world settings.
\end{itemize}

\section{Related Work}

\subsection{Vision-and-Language Navigation}
Vision-and-Language Navigation (VLN) studies how an embodied agent executes natural-language instructions within complex, visually grounded environments~\cite{vln}. Early research focused on discrete graph-based formulations~\cite{r2r,Matterport3D}, which have since evolved toward more realistic continuous settings~\cite{r2r_ce,Habitat}. A predominant trend in the field emphasizes enhancing the agent's cross-modal understanding as the primary path to performance gains. Consequently, recent state-of-the-art methods have leveraged increasingly powerful pre-trained backbones~\cite{DualVLN,PROSPECT,DyGeoVLN,internvla-n1}, massive-scale navigation and auxiliary datasets~\cite{NavFoM,StreamVLN,Uni-NaVid,NaVILA}, and sophisticated temporal modeling~\cite{JanusVLN,NavForesee,AstraNav-World,SPAN-Nav}. While these advancements have significantly pushed the performance frontier, it is increasingly recognized that understanding ability is not the sole determinant of success in VLN. Recent studies have begun to highlight critical external factors that pose significant challenges to even the most capable agents. Specifically, natural-language instructions often contain inherent ambiguities~\cite{mvlfvln,wotevlnwmmp}, and agents frequently exhibit sensitivity to visual disturbances or environmental shifts~\cite{dtebivln,rtegivlnahsopavd}.
Despite their importance, these factors—instruction ambiguity and visual robustness—are often treated as anecdotal observations rather than being systematically analyzed. Their individual and joint impacts on navigation performance remain insufficiently quantified. Motivated by this, we conduct analysis studies to show how these factors constrain current VLN agents, aiming to provide a more nuanced understanding of the remaining bottlenecks in the field.

\subsection{Vision-Language-Action Models for VLN}
Recent advances in large-scale Vision-Language Models (VLMs)~\cite{Qwen2_5_VL,LLaMA_VID,LLaVA_Video,VILA,Vicuna} have inspired a growing body of work that adapts them to Vision-and-Language Navigation. A representative direction is to fine-tune pre-trained VLMs to directly map observations and language instructions to low-level navigation actions in an end-to-end manner, forming Vision-Language-Action models for VLN~\cite{NaVid,Uni-NaVid,NaVILA,StreamVLN,CorrectNav,JanusVLN}. Benefiting from the broad visual-semantic priors learned from large-scale internet data, this paradigm provides a promising route for policy learning in complex environments~\cite{pboupatlvsim}. Early efforts such as NaVid~\cite{NaVid} demonstrated the feasibility of monocular RGB-based end-to-end navigation, while later methods such as StreamVLN~\cite{StreamVLN} further improved performance with stronger foundation models and larger-scale training data. However, existing methods still suffer from degraded navigation performance and reduced execution consistency in realistic deployment settings In particular, their sensitivity to visual disturbances and instruction ambiguity often results in unstable cross-scene transfer and unsatisfactory accuracy, even in relatively controlled settings~\cite{vlntatasiteofm,vln}. Motivated by these limitations, we introduce StereoNav, which leverages clear or fuzzy target-point priors to mitigate instruction ambiguity, while combining stereo-based unified understanding and joint prediction modeling to improve robustness under visual perturbations and enhance navigation accuracy.

\begin{figure*}[!b]
    \centering
    \includegraphics[width=\linewidth]{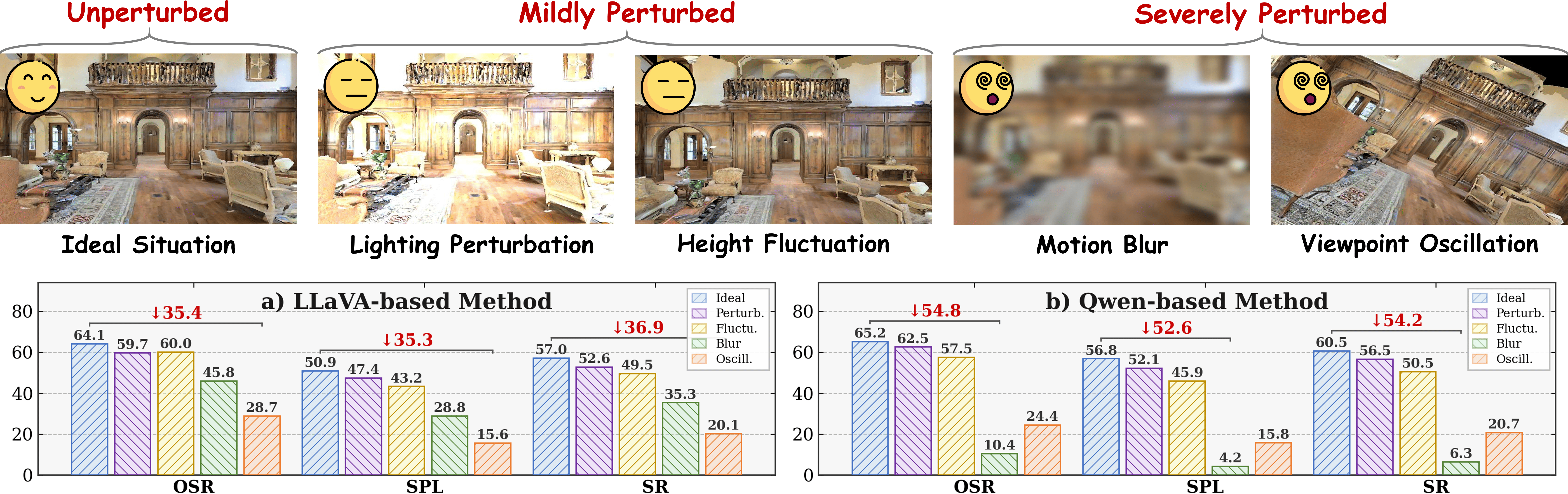}
    \vspace{-1.0em}
    \caption{\textbf{Impact of visual uncertainty on VLN agents.} (a) \textbf{Top:} Visual examples of four common perturbations during embodied navigation. (b) \textbf{Bottom:} Performance degradation of representative open-source VLN methods, where the LLaVA-based and Qwen-based methods correspond to StreamVLN~\cite{StreamVLN} and JanusVLN~\cite{JanusVLN}, respectively. Although existing agents perform competitively in the ideal setting, their navigation performance degrades markedly under severe visual perturbations.}
    \vspace{-1.0em}
    \label{fig:visual_uncertainty}
\end{figure*}

\section{Revisiting the Limitations of VLN}

\label{sec:Pilot_Study}
To understand the performance degradation in physical VLN deployments, we identify two primary bottlenecks: perception instability (\ref{sec:visual_uncertainty}) and instructional under-specification (\ref{sec:instruction_ambiguity}). We analyze their impact below to provide the empirical motivation for our design.

% We systematically investigate why VLN methods often suffer from degraded navigation performance and reduced execution consistency when deployed in physical environments. Our analysis shows that this limitation mainly arises from two fundamental factors: visual uncertainty~\ref{sec:visual_uncertainty} and the ambiguity of instructional guidance~\ref{sec:instruction_ambiguity}. These factors jointly hinder the robustness and practicality of VLN. In this section, we analyze them in detail to motivate the subsequent design.

\subsection{The Impact of Visual Uncertainty}
\label{sec:visual_uncertainty}

While VLN methods excel in in-domain evaluations, their performance often degrades in real-world deployments due to perception instability. As shown in Figure~\ref{fig:visual_uncertainty}, this instability commonly manifests as: lighting perturbation, height fluctuation, motion blur, and viewpoint oscillation~\cite{ASOTROCVMACC, RobustNav}. We categorize the former two as mild perturbations and the latter as severe disturbances. Through pilot studies (see Appendix~\ref{sec:Visual_Uncertainty_Experimental_Setup}), we systematically examine how these factors constrain agent performance to motivate our design.

% Prior work~\cite{NaVid, Uni-NaVid, RethinkingGap, StreamVLN} has shown that although many VLN methods achieve competitive results in in-domain evaluations, their performance degrades notably across environments, especially in real-world deployment. Since visual observation is the primary basis for an embodied agent's perception and navigation decisions, we investigate whether this gap arises, beyond model understanding itself, from visual uncertainty. As shown in Figure~\ref{fig:visual_uncertainty}, existing studies~\cite{ASOTROCVMACC, RobustNav} suggest four common forms of visual uncertainty, namely \textit{lighting perturbation}, \textit{height fluctuation}, \textit{motion blur}, and \textit{viewpoint oscillation}. The first two are treated as mild perturbations, while the latter two are regarded as severe disturbances. To examine their effects, we conduct a series of pilot studies to better understand how visual uncertainty influences agent performance. Details are provided in Appendix~\ref{sec:Visual_Uncertainty_Experimental_Setup}.

As shown in Fig.~\ref{fig:visual_uncertainty}, current VLN agents exhibit a significant lack of robustness to perception instability. Under mild perturbations, including lighting changes and height fluctuations, the degradation remains relatively contained, performance degradation remains relatively contained, with key metrics like Success Rate (SR) and Success weighted by Path Length (SPL) typically dropping by around 10\% compared to the ideal situation. However, severe disturbances lead to a dramatic performance collapse. For the LLaVA-based method, SR plummets from 57.0\% to 35.3\% under motion blur, and further to 20.1\% under viewpoint oscillation. The Qwen-based method proves even more vulnerable; notably, its SPL drops from 56.8\% to a mere 4.2\% under motion blur—less than 8\% of its original performance. In real-world deployments, changes in visual perception due to environmental variations (such as lighting disturbances) are very common. However, existing VLN models are extremely sensitive to such changes, necessitating a solution to ensure reliable practical applications.

%These sharp declines across all metrics, especially in high-dynamic scenarios, underscore that the primary bottleneck in realistic deployment lies not just in high-level reasoning, but in the extreme sensitivity of the underlying visual perception to embodied disturbances.

%StreamVLN shows an average reduction of more than 25\% across all metrics under severe perturbations, while JanusVLN is affected even more substantially. Notably, under motion blur, all metrics of JanusVLN fall to below 16\% of their counterparts in the ideal setting. These results indicate that the weakness of current VLN agents lies not only in high-level understanding, but also in the vulnerability of visual perception under realistic embodied disturbances.

\begin{figure}[htbp]
    \centering
    \includegraphics[width=\columnwidth]{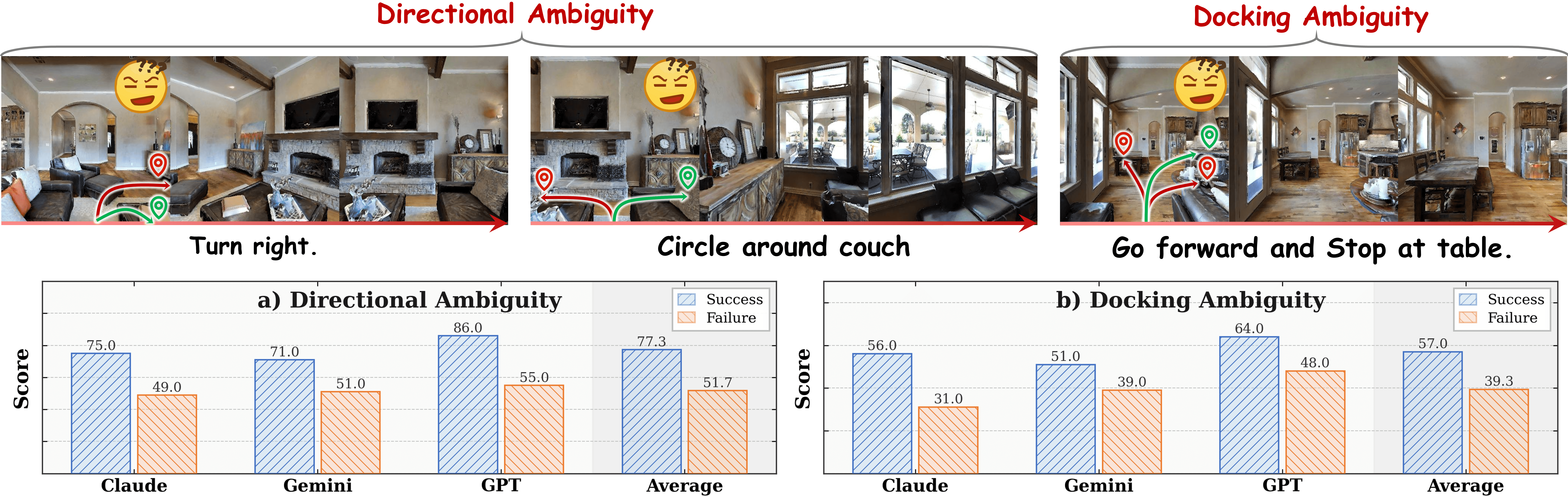}
    \vspace{-1.0em}
    \caption{\textbf{Impact of instructional under-specification on VLN agents.}
    (a) \textbf{Top:} Representative cases of Directional Ambiguity, where under-specified route or orientation cues permit multiple feasible paths, and Docking Ambiguity, where vague goal descriptions permit multiple plausible stopping targets.
    (b) \textbf{Bottom:} Distributions of ambiguity scores across representative VLMs, where lower scores indicate stronger ambiguity effects.}
    \vspace{-1.0em}
    \label{fig:instruction_ambiguity}
\end{figure}

\subsection{The Impact of Instruction Ambiguity}
\label{sec:instruction_ambiguity}

%Instructions are typically provided by humans to guide an agent through an environment, yet they are often inherently ambiguous and can lead to misinterpretation. Since instructions are the primary guidance for VLN agents, we study how such ambiguity affects navigation performance. 

Human-provided instructions serve as the primary guidance for VLN, yet their inherent instructional under-specification often creates a challenging one-to-many mapping between a single utterance and multiple plausible navigation paths. As shown in Figure~\ref{fig:instruction_ambiguity}, insufficient instruction description manifests as two different types of decision uncertainty: \textit{Directional Ambiguity}, which complicates path selection during intermediate navigation phases; and \textit{Docking Ambiguity}, which disrupts target localization at termination. Our cross-model analysis reveals a strong correlation between the degree of under-specification and navigation failure, identifying these ambiguities as fundamental failure modes (detailed in Appendix~\ref{sec:Instruction_Ambiguity_Experimental_Setup}).

%As shown in Fig.~\ref{fig:instruction_ambiguity}, vague instructions mainly induce two types of ambiguity: \textit{Directional Ambiguity}, where multiple plausible paths or orientations exist during navigation, and \textit{Docking Ambiguity}, where multiple stopping locations or target objects are plausible near termination. We conduct pilot studies to analyze their relationship with navigation outcomes, with details provided in Appendix~\ref{sec:Instruction_Ambiguity_Experimental_Setup}.

% Instructions are typically provided by humans to guide an agent through an environment. Yet natural language instructions are often inherently ambiguous, which can easily cause misinterpretation. Since instructions serve as the sole source of guidance for a VLN agent, a natural question arises: how much does instruction ambiguity affect navigation performance? As shown in Figure~\ref{fig:instruction_ambiguity}, ambiguity induced by vague instructions mainly appears in two forms. One is \textit{Directional Ambiguity}, which arises during navigation when the instruction admits multiple plausible paths or orientations. The other is \textit{Docking Ambiguity}, which arises at the stopping stage when the instruction admits multiple plausible stopping locations or target objects. To better understand how instruction ambiguity affects navigation performance, we conduct a series of pilot studies to investigate its relationship with navigation outcomes. Details are provided in Appendix~\ref{sec:Instruction_Ambiguity_Experimental_Setup}.

As shown in Fig.~\ref{fig:instruction_ambiguity}, instruction ambiguity is strongly associated with navigation failure, where lower scores indicate a more severe influence from the corresponding ambiguity type. Specifically, failed cases are more strongly affected by directional ambiguity than successful ones, with average scores of 51.7 and 77.3, respectively, indicating that under-specified path or orientation cues can easily mislead the agent onto an unintended route, thereby causing task failure. Docking ambiguity is also evident. Failed cases score 17.7 points lower than successful ones. This indicates that docking ambiguity is widespread in VLN instructions, while failed cases are often affected by more severe ambiguity around the stopping location or target object. Overall, these results show that instruction ambiguity directly increases the likelihood of navigation failure.

% Fig.~\ref{fig:instruction_ambiguity} shows that instruction ambiguity is strongly associated with navigation failure, where lower scores indicate more severe influence from the corresponding ambiguity type. Failed cases are more affected by directional ambiguity than successful ones, with average scores of 51.7 and 77.3, respectively, suggesting that under-specified path or orientation cues can mislead agents onto unintended routes. Docking ambiguity is also evident: failed cases score 17.7 points lower than successful ones, indicating that ambiguity around stopping locations or target objects further increases the risk of failure. Overall, these results show that instruction ambiguity directly undermines reliable navigation.

\section{Methodology}
In this section, we first define the task in Sec.~\ref{sec:problem_formulation}, and then introduce \textit{StereoNav}. The proposed method alleviates instruction under-specification by incorporating precise or coarse target-point priors (Sec.~\ref{sec:visual_rendering}), while leveraging stereo-based unified understanding modeling (Sec.~\ref{sec:unified_understanding_modeling}) and joint prediction modeling (Sec.~\ref{sec:joint_prediction_modeling}) to improve robustness against visual disturbances and achieve more accurate navigation.

\begin{figure}[htbp]
    \centering
    \includegraphics[width=\linewidth]{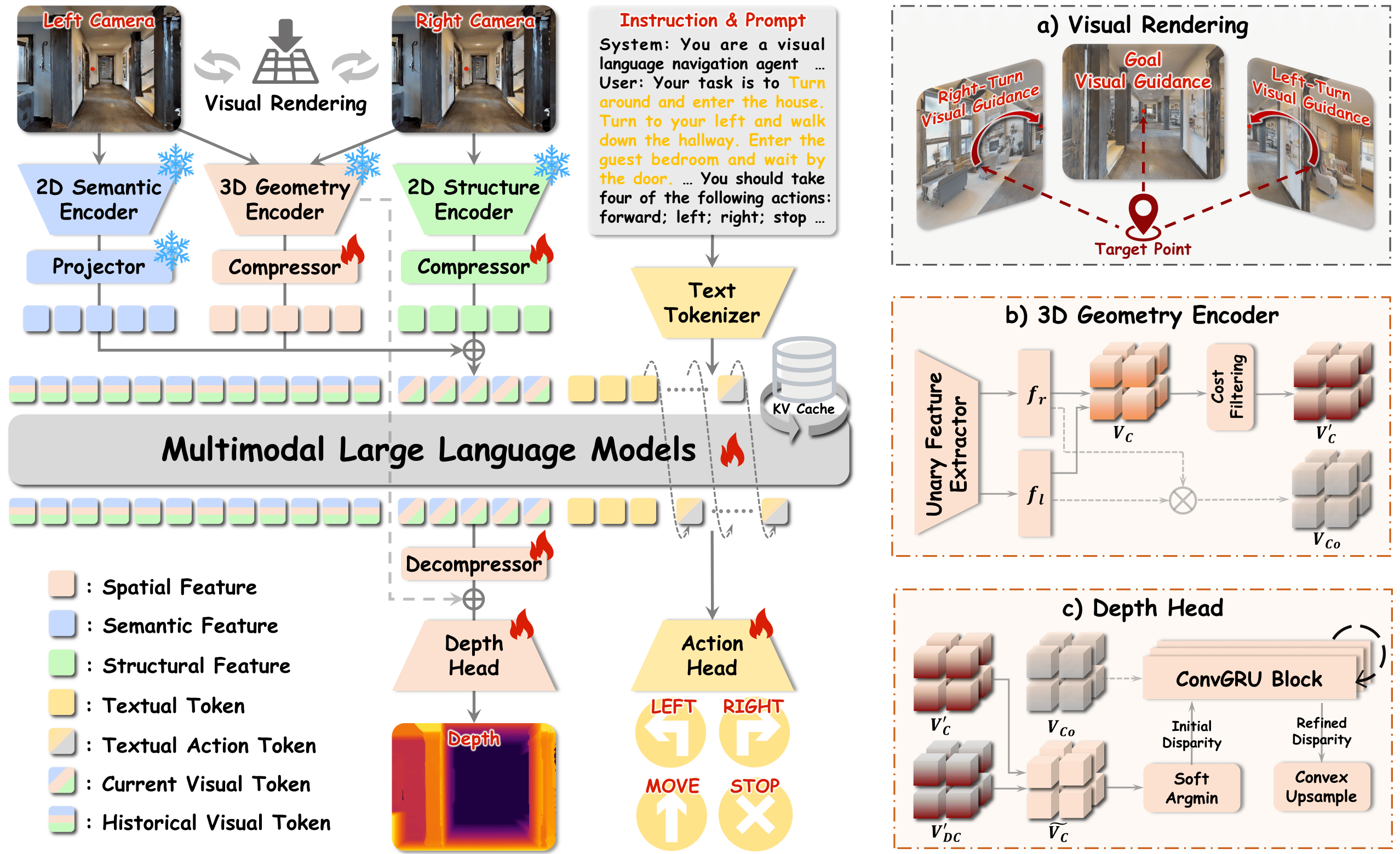}
    \caption{\textbf{Overview of StereoNav.}
StereoNav takes stereo RGB observations, a navigation instruction, and a target-location prior as input. The target prior is rendered as persistent visual guidance, while stereo observations are encoded into unified semantic, structural, and geometric tokens through 2D semantic, 2D structural, and 3D geometry encoders. These tokens are then processed by the MLLM for joint action and depth prediction. The right panels illustrate detailed designs of selected modules.}
    \label{fig:main_figure}
\vspace{-0.5cm}
\end{figure}

\subsection{Problem Formulation}
\label{sec:problem_formulation}

We consider Vision-and-Language Navigation in Continuous 3D Environments. In our setting, the agent is given a natural-language instruction \(I\) and a fuzzy or precise target-location cue \(p\) for global guidance. At each time step \(t\), it receives stereo RGB observations \(o_t^{l}\) and \(o_t^{r}\), together with the history of past stereo observations \(\mathcal{H}_{t-1}=\{(o_1^{l},o_1^{r}), \dots, (o_{t-1}^{l},o_{t-1}^{r})\}\). Based on these inputs, the agent jointly predicts a navigation action \(a_t\) and a dense depth estimate \(\hat{d}_t\):
\begin{equation}
(a_t, \hat{d}_t) = \pi(o_t^{l}, o_t^{r}, I, p, \mathcal{H}_{t-1}), \quad t = 1,2,\dots,T.
\end{equation}
The action space is \(\mathcal{A}=\{\texttt{move forward}, \texttt{turn left}, \texttt{turn right}, \texttt{stop here}\}\), corresponding to a forward translation of 0.25 m, \(15^\circ\) left/right rotations, and episode termination, respectively. Action prediction is the primary objective, while depth prediction serves as an auxiliary geometric objective for improving spatial perception, robustness, and navigation accuracy.

\subsection{Visual Rendering}
\label{sec:visual_rendering}
% Existing VLN methods primarily rely on natural-language instructions for navigation~\cite{NaVid,Uni-NaVid,NaVILA,StreamVLN}. However, as discussed in Sec.~\ref{sec:Pilot_Study}, such instructions are often inherently ambiguous, which can easily cause the agent to deviate from the intended route and lead to navigation failure. Our key insight is: \textit{in real human--robot interaction, the speaker typically has coarse or precise prior knowledge of the goal location when giving navigation instructions.} Motivated by this observation, we explicitly exploit such a spatial prior. Unlike prior point-goal navigation methods~\cite{NoMaD,NavDP} that directly introduce target-location information as a symbolic input, we transform the target prior into a visual guidance signal embedded in the agent's own observations, thereby facilitating its effective use during navigation. 
Existing VLN agents rely heavily on natural-language instructions~\cite{NaVid,Uni-NaVid,NaVILA,StreamVLN}, which are inherently prone to instructional under-specification (Sec.~\ref{sec:Pilot_Study}) and often lead to navigation failure. We observe that in practical human--robot interaction, speakers typically possess varying degrees of prior knowledge regarding the goal's location. Motivated by this, we explicitly leverage such spatial priors. Unlike conventional point-goal methods~\cite{NoMaD,NavDP} that treat target coordinates as symbolic inputs, we project the spatial prior directly into the agent’s egocentric visual space, transforming abstract coordinates into intuitive visual guidance that facilitates more effective decision-making.

As illustrated in Fig.~\ref{fig:main_figure}a, the agent is initialized with both a language instruction and a target-location prior (ranging from fuzzy to precise). This prior is dynamically updated based on the current pose and rendered onto the stereo observations: visible targets are marked with a red point, while out-of-view targets are indicated by boundary-aware cues. This design provides a lightweight yet persistent global directional signal, grounding the agent's target awareness while preserving its capacity for fine-grained local action prediction. By bridging the gap left by ambiguous instructions, this mechanism mitigates both Directional and Docking Ambiguity, thereby enhancing navigation robustness.

% As shown in Fig.~\ref{fig:main_figure} a), the agent receives both the instruction and a fuzzy or precise target location prior to initialization. At each time step, the prior is updated under the current pose and rendered onto the stereo observations: visible targets are marked by a red point, while out-of-view targets are indicated by a boundary-aware cue at the left or right image boundary. Further details are provided in Appendix~\ref{sec:Appendix_Visual_Rendering}. This design provides the agent with a lightweight yet persistent global directional prior, enabling target awareness while preserving fine-grained local action prediction. By compensating for ambiguous language guidance, it alleviates the effects of Directional Ambiguity and Docking Ambiguity, thereby improving navigation robustness and accuracy. 

\subsection{Unified Understanding Modeling}
\label{sec:unified_understanding_modeling}

Although existing end-to-end VLN methods~\cite{StreamVLN,NaVid,NavFoM} commonly inherit the visual encoder of the underlying MLLM, these representations are mainly optimized for cross-modal semantic alignment rather than the structural and geometric perception required for embodied navigation. This makes them vulnerable to visual perturbations and can undermine reliable navigation. To address this limitation, we propose Unified Understanding Modeling, which leverages stereo observations to jointly capture semantic, structural, and geometric cues. By exploiting their complementarity and binocular redundancy, our design improves perceptual stability and supports more robust navigation.

To achieve Unified Understanding Modeling, we construct a multimodal 2D--3D representation that synergizes semantic, structural, and geometric cues. For 2D features, we employ InternViT to maintain semantic grounding from large-scale pretraining and DINOv2 to capture fine-grained structural patterns like layouts and contours. Applying these encoders to stereo views inherently preserves binocular disparity, enhancing the agent's spatial discrimination. To incorporate explicit depth priors, we introduce a 3D geometry encoder inspired by FoundationStereo (Fig.~\ref{fig:main_figure}b). This module processes stereo pairs into a 4D cost volume $V_c$, which is refined through attentive hybrid filtering ($V_c'$) and tokenized for fusion. The resulting unified tokens $T_t$ are formulated as:\begin{equation}T_t = \alpha P(F_t^{\mathrm{sem}}) + \beta C(F_t^{\mathrm{str}}) + \gamma G(V_{c,t}^{\prime}),\end{equation}where $F_t^{\mathrm{sem}}$ and $F_t^{\mathrm{str}}$ represent the semantic and structural features, while $P(\cdot)$, $C(\cdot)$, and $G(\cdot)$ denote their respective projection and tokenization modules. $\alpha$, $\beta$, and $\gamma$ are learnable fusion weights. This integration yields a compact yet comprehensive representation, ensuring robust perception for embodied navigation.

\subsection{Joint Prediction Modeling}
\label{sec:joint_prediction_modeling}
Previous studies suggest that explicit depth guidance~\cite{Dynam3D,AgentVLN,NaVid-4D} and auxiliary future observation prediction~\cite{NavForesee,SPAN-Nav,NavMorph} can improve navigation and representation learning. However, these designs remain costly and only partially effective for robust embodied navigation: depth sensing increases hardware dependency and is vulnerable to environmental interference, while future prediction introduces extra computation and mainly provides indirect supervision. To address these limitations, we propose joint action-depth prediction, where depth prediction serves as geometric supervision for implicit 3D perception and further supports more robust action decisions.

Concretely, we instantiate this objective with two lightweight prediction heads: an action head for future action generation and a depth head for stereo disparity estimation. For action prediction, we follow prior VLM-based navigation methods~\cite{NaVid,StreamVLN}. At each time step, the model autoregressively generates the textualized action sequence for the next four steps, conditioned on the system prompt, historical observations, the current observation, and the navigation instruction. The detailed prompt template used for action generation is provided in Appendix~\ref{sec:Prompt_Template}:
\begin{equation}
x =
\underbrace{[S_{\mathrm{sys}}]\, \tilde{s} \,[E_{\mathrm{sys}}] \,
[S_{\mathrm{usr}}]\, \tilde{o}_{1:t-1},~\tilde{o}_t,~\tilde{l} \,[E_{\mathrm{usr}}]}_{\text{Input}}
\;
\underbrace{[S_{\mathrm{asst}}]\, \tilde{a}_{t:t+3} \,[E_{\mathrm{asst}}]}_{\text{Generation}},
\end{equation}
where \(S_{*}\) and \(E_{*}\) denote the segment boundary tokens, and \(\tilde{a}_{t:t+3}\) denotes the generated four-step action sequence. Meanwhile, the depth head adopts a coarse-to-fine stereo matching strategy, as shown in Fig.~\ref{fig:main_figure} c). It takes the filtered stereo cost volume \(V'_C\) from the 3D Geometry Encoder and the decoded cost volume \(V'_{DC}\) reconstructed from current-view tokens, and estimates the final disparity through cost-volume fusion, Soft Argmin initialization, recurrent refinement, and convex upsampling:
\begin{equation}
\hat{d} =
\mathrm{Upsample}\big(
\mathrm{ConvGRU}\big(
\sum\nolimits_{\delta=0}^{D/4-1}
\delta \cdot \mathrm{Softmax}(V'_C + V'_{DC})(\delta),
V_{co}
\big)
\big),
\end{equation}
Here, \(V_{co}\) denotes the geometry-aware contextual feature from the 3D Geometry Encoder, which guides ConvGRU-based disparity refinement.

\begin{figure*}[!t]
    \centering
    \includegraphics[width=\linewidth]{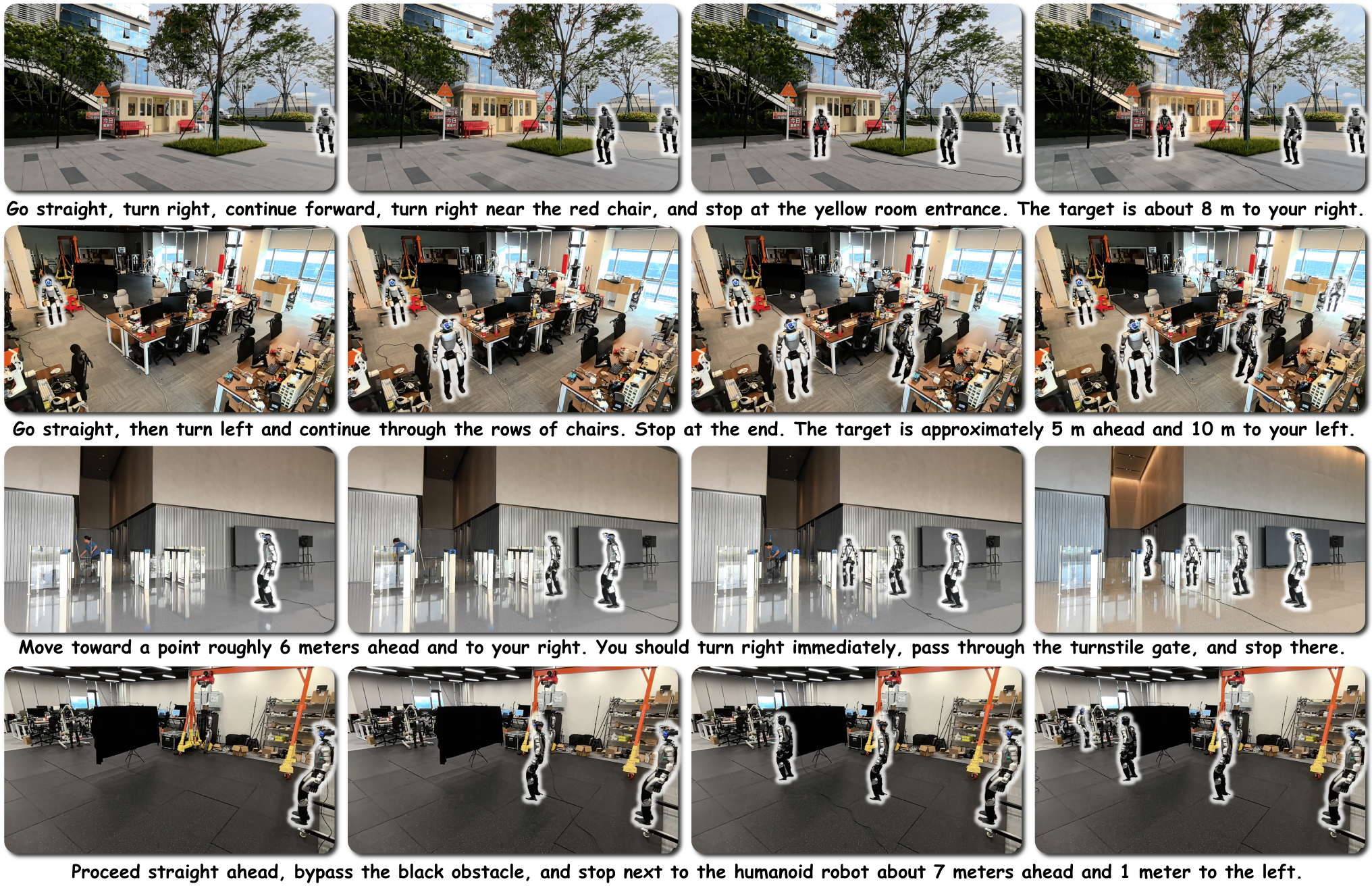}
    \vspace{-1.1em}
    \caption{\textbf{Qualitative examples of StereoNav in real-world.} From top to bottom: Outdoor, Office, Lobby, and Gym. The results demonstrate StereoNav’s reliability across diverse scenes. Note that these examples are visualized from a third-person perspective; details regarding the actual sensor inputs used for navigation are provided in Section~\ref{sec:real_world_evaluation}.}
    \vspace{-1.0em}
    \label{fig:real_world}
\end{figure*}

\section{Experiment}
We evaluate StereoNav from multiple perspectives. We first present the experimental setup in Sec.~\ref{sec:experimental_details}; then report comprehensive comparisons in simulated and real-world environments in Sec.~\ref{sec:comprehensive_performance_evaluation}; analyze robustness and reliability in Sec.~\ref{sec:robustness_reliability_evaluation}; and provide ablation studies in Sec.~\ref{sec:ablation_study}.

% In this section, we evaluate StereoNav from multiple perspectives. We first present the implementation details, benchmarks, and real-world deployment protocol in Sec.~\ref{sec:experimental_details}. We then provide comprehensive comparisons in simulated and real-world environments in Sec.~\ref{sec:comprehensive_performance_evaluation}, followed by robustness and reliability evaluations under deployment-oriented conditions in Sec.~\ref{sec:robustness_reliability_evaluation}. Finally, we conduct ablation studies in Sec.~\ref{sec:ablation_study} to analyze the contribution of each key component in StereoNav.

\subsection{Experimental Details}
\label{sec:experimental_details}

\begin{table}[t]
    \centering
    \caption{\textbf{Comparison with SOTA methods on the R2R/RxR-CE Val-Unseen.} Light and dark blue rows indicate StereoNav trained with standard navigation data only and with additional external data, respectively. Top-three results are marked by \textbf{bold}, \underline{underline}, and \dotuline{dotted underline}. Parentheses report the signed difference from the best metric value among Egocentric RGB Agent baselines.}
    \resizebox{\columnwidth}{!}{
    \begin{tabular}{cc|ccc|cc|cccc|ccc}
        \toprule
        \multirow{2}{*}{\textbf{Method}} &
        \multirow{2}{*}{\textbf{Size}} &
        \multicolumn{3}{c|}{\textbf{Observation}} &
        \multicolumn{2}{c|}{\textbf{Prediction}} & 
        \multicolumn{4}{c|}{\textbf{R2R-CE Val-Unseen}} & 
        \multicolumn{3}{c}{\textbf{RXR-CE Val-Unseen}} \\
        & & \textbf{Pano.} & \textbf{Depth} & \textbf{RGB} & \textbf{Extra} & \textbf{Action} & \textbf{NE$\downarrow$} & \textbf{OSR$\uparrow$} & \textbf{SR$\uparrow$} & \textbf{SPL$\uparrow$} & \textbf{NE$\downarrow$} & \textbf{SR$\uparrow$} & \textbf{SPL$\uparrow$} \\
        \midrule\midrule

        \midrule
        \rowcolor{gray!15}\multicolumn{14}{c}{\textit{Panoramic RGB-D Agent}} \\

        % 2024-1-22
        ETPNav~\cite{ETPNav} & - & \ding{51} & \ding{51} & \ding{55} & \ding{55} & \ding{51} & 4.7 & 65.0 & 57.0 & 49.0 & 5.6 & 54.8 & 44.9 \\

        % 2025-12-11
        CLASH~\cite{CLASH} & - & \ding{51} & \ding{51} & \ding{55} & \ding{55} & \ding{51} & 4.1 & 73.0 & 65.0 & 55.0 & - & - & - \\

        % 2025-12-14
        D3D-VLP~\cite{D3D-VLP} & 2B & \ding{51} & \ding{51} & \ding{55} & \ding{55} & \ding{51} & 4.7 & 67.2 & 61.3 & 56.1 & - & - & - \\

        % 2025-12-24
        ETP-R1~\cite{ETP-R1} & - & \ding{51} & \ding{51} & \ding{55} & \ding{55} & \ding{51} & 3.9 & 72.0 & 65.0 & 56.0 & 5.2 & 59.9 & 49.0 \\

        % 2026-03-18
        P$^3$Nav~\cite{P-Nav} & - & \ding{51} & \ding{51} & \ding{55} & \ding{55} & \ding{51} & 4.4 & 69.0 & 62.0 & 52.0 & 5.4 & 58.0 & 47.9 \\

        \midrule
        \rowcolor{gray!15}\multicolumn{14}{c}{\textit{Panoramic RGB Agent}} \\

        % 2024-8-20
        AO-Planner~\cite{AO-Planner} & - & \ding{51} & \ding{55} & \ding{55} & \ding{55} & \ding{51} & 5.6 & 59.0 & 47.0 & 33.0 & 7.1 & 43.3 & 30.5 \\

        % 2025-09-16
        NavFoM~\cite{NavFoM} & 7B & \ding{51} & \ding{55} & \ding{55} & \ding{55} & \ding{51} & 4.6 & 72.1 & 61.7 & 55.3 & 4.7 & 64.4 & 56.2 \\

        % 2026-02-12
        ABot-N0~\cite{ABot-N0} & 4B & \ding{51} & \ding{55} & \ding{55} & \ding{55} & \ding{51} & \dotuline{3.8} & 70.8 & 66.4 &  \dotuline{63.9} & \textbf{3.8} & \dotuline{69.3} & \dotuline{60.0} \\

        % 预测 world 信息
        % 2026-03-13
        NavForesee~\cite{NavForesee} & 3B & \ding{51} & \ding{55} & \ding{55} & \ding{51} & \ding{51} & 3.9 &  \underline{78.4} & 66.2 & 59.7 & \dotuline{4.2} & 66.3 & 53.2 \\

        % 预测 world 信息
        % 2026-03-16
        SPAN-Nav~\cite{SPAN-Nav} & - & \ding{51} & \ding{55} & \ding{55} & \ding{51} & \ding{51} & 4.1 & 75.3 & 66.3 & 59.3 & \dotuline{4.2} & \textbf{69.7} & \underline{60.1} \\

        \midrule
        \rowcolor{gray!15}\multicolumn{14}{c}{\textit{Egocentric RGB-D Agent}} \\
        % 2024
        NaVid-4D~\cite{NaVid-4D} & 7B & \ding{55} & \ding{51} & \ding{51} & \ding{55} & \ding{51} & 6.0 & 55.7 & 43.8 & 37.1 & - & - & - \\
        
        % 2025-05-16
        Dynam3D~\cite{Dynam3D} & 7B & \ding{55} & \ding{51} & \ding{51} & \ding{55} & \ding{51} & 5.3 & 62.1 & 52.9 & 45.7 & - & - & - \\

        % 2025-07-22
        NavMorph~\cite{NavMorph} & - & \ding{55} & \ding{51} & \ding{51} & \ding{51} & \ding{51} & 5.8 & 56.9 & 47.9 & 33.2 & 8.9 & 30.8 & 22.8 \\
        
        % 2025-09-11
        InternVLA-N1~\cite{internvla-n1} & 8B & \ding{55} & \ding{51} & \ding{51} & \ding{55} & \ding{51} & 4.8 & 63.3 & 58.2 & 54.0 & 5.9 & 53.5 & 46.1 \\

        % 2026-03-18
        AgentVLN~\cite{AgentVLN} & 3B & \ding{55} & \ding{51} & \ding{51} & \ding{55} & \ding{51} & 3.9 & 73.5 & \dotuline{67.2} & \underline{64.7} & \underline{3.9} & \underline{69.5} & \textbf{61.3} \\

        \midrule
        \rowcolor{gray!15}\multicolumn{14}{c}{\textit{Egocentric RGB Agent}} \\
        
        % 2024-01-30
        NaVid~\cite{NaVid} & 7B & \ding{55} & \ding{55} & \ding{51} & \ding{55} & \ding{51} & 5.5 & 49.1 & 37.4 & 35.9 & - & - & - \\

        % 2024-12-09
        Uni-NaVid~\cite{Uni-NaVid} & 7B & \ding{55} & \ding{55} & \ding{51} & \ding{55} & \ding{51} & 5.6 & 53.3 & 47.0 & 42.7 & 6.2 & 48.7 & 40.9 \\ % OSR:  55.5

        % 2025-02-17
        NaVILA~\cite{NaVILA} & 8B & \ding{55} & \ding{55} & \ding{51} & \ding{55} & \ding{51} & 5.2 & 62.5 & 54.0 & 49.0 & 6.8 & 49.3 & 44.0 \\

        % 2025-07-07
        StreamVLN~\cite{StreamVLN} & 7B & \ding{55} & \ding{55} & \ding{51} & \ding{55} & \ding{51} & 5.0 & 64.2 & 56.9 & 51.9 & 6.2 & 52.9 & 46.0 \\

        % 2025-09-11
        InternVLA-N1~\cite{internvla-n1} & 8B & \ding{55} & \ding{55} & \ding{51} & \ding{55} & \ding{51} & 4.9 & 60.6 & 55.4 & 52.1 & 6.4 & 49.5 & 41.8 \\
        
        % 2025-09-16
        NavFoM~\cite{NavFoM} & 7B & \ding{55} & \ding{55} & \ding{51} & \ding{55} & \ding{51} & 5.0 & 64.9 & 56.2 & 51.2 & 5.5 & 57.4 & 49.4 \\

        % 2025-12-09
        DualVLN~\cite{DualVLN} & 8B & \ding{55} & \ding{55} & \ding{51} & \ding{55} & \ding{51} & 4.1 & 70.7 & 64.3 & 58.5 & 4.6 & 61.4 & 51.8 \\

        % 2025-12-11
        Efficient-VLN~\cite{Efficient-VLN} & 4B & \ding{55} & \ding{55} & \ding{51} & \ding{55} & \ding{51} & 4.2 & 73.7 & 64.2 & 55.9 & \underline{3.9} & 67.0 & 54.3 \\

        % 2026-02-25
        JanusVLN~\cite{JanusVLN} & 8B & \ding{55} & \ding{55} & \ding{51} & \ding{55} & \ding{51} & 4.8 & 65.2 & 60.5 & 56.8 & 6.1 & 56.2 & 47.5 \\

        % 预测 world 信息
        % 2026-03-04
        PROSPECT~\cite{PROSPECT} & 9B & \ding{55} & \ding{55} & \ding{51} & \ding{51} & \ding{51} & 4.9 & 65.2 & 58.9 & 54.0 & 5.7 & 54.6 & 46.2 \\
        
        % 2026-03-10
        SACA~\cite{SACA} & 8B & \ding{55} & \ding{55} & \ding{51} & \ding{55} & \ding{51} & 4.2 &  69.3 &  64.7 & 56.9 & 4.8 & 62.1 & 51.7 \\

        % 2026-03-16
        NaVIDA~\cite{NaVIDA} & 3B & \ding{55} & \ding{55} & \ding{51} & \ding{55} & \ding{51} & 4.3 & 69.5 & 61.4 & 54.7 & 5.2 & 57.4 & 49.6 \\

        % 预测 world 信息
        % 2026-03-22
        DyGeoVLN~\cite{DyGeoVLN} & 9B & \ding{55} & \ding{55} & \ding{51} & \ding{51} & \ding{51} & 4.4 & 70.1 & 60.8 & 55.8 & - & - & - \\

        % 2026-03-26
        DecoVLN~\cite{DecoVLN} & 7B & \ding{55} & \ding{55} & \ding{51} & \ding{55} & \ding{51} & 5.0 & 63.5 & 56.3 & 50.5 & 5.7 & 54.2 & 46.3 \\

        \rowcolor{blue!4}
        StereoNav & 3B & \ding{55} & \ding{55} & \ding{51} & \ding{51} & \ding{51} & \underline{3.0}~\textbf{(-1.1)} & \dotuline{76.6}~\textbf{(+2.9)} & \underline{72.8}~\textbf{(+8.1)} & 56.4~\textbf{(+2.1)} & 5.9~\textbf{(+2.0)} & 58.0~\textbf{(-9.0)} & 43.5~\textbf{(-8.5)} \\

        \rowcolor{blue!8}
        StereoNav & 3B & \ding{55} & \ding{55} & \ding{51} & \ding{51} & \ding{51} & \textbf{2.1}~\textbf{(-2.0)} & \textbf{82.4}~\textbf{(+8.7)} & \textbf{81.1}~\textbf{(+16.4)} & \textbf{68.3}~\textbf{(+9.8)} & 4.6~\textbf{(+0.7)} & 67.5~\textbf{(+0.5)} & 52.0~\textbf{(-2.3)} \\
        
        \bottomrule
    \end{tabular}}
    \vspace{-1.0em}
    \label{tab:main}
\end{table}

\textbf{Implementation Details.}
StereoNav is built upon InternVL-3.5-2B~\cite{InternVL3_5}. The 2D semantic encoder is initialized from the InternViT visual encoder of InternVL-3.5, while the 2D structure encoder is initialized from DINOv2~\cite{DINOv2}; both the 3D geometry encoder and the depth head are initialized from FoundationStereo~\cite{FoundationStereo}. During training, stereo images are resized to \(448 \times 448\), and 8 historical frames are uniformly sampled as visual context. StereoNav autoregressively predicts the next 4 navigation actions and is trained following a two-stage strategy, with the 2D semantic, 2D structure, and 3D geometry branches fused using weights of 1.0, 0.6, and 0.2, respectively. Additional implementation details are provided in Appendix~\ref{sec:Training_Details_and_Configurations}.

% StereoNav is built upon InternVL-3.5-2B~\cite{InternVL3_5}. The 2D semantic encoder is initialized from the InternViT visual encoder of InternVL-3.5, while the 2D structure encoder is initialized from DINOv2~\cite{DINOv2}; both the 3D geometry encoder and the depth head are initialized from FoundationStereo~\cite{FoundationStereo}. During training, stereo images are resized to \(448 \times 448\), and 8 historical frames are used as visual context. StereoNav autoregressively predicts the next 4 navigation actions and is trained following a two-stage strategy. We use a base learning rate of \(2.0\times10^{-5}\), with a smaller learning rate of \(2.0\times10^{-6}\) for the 2D structure encoder and the depth head. During feature fusion, the 2D semantic, 2D structure, and 3D geometry branches are weighted by 1.0, 0.6, and 0.2, respectively. Additional training details and configurations are provided in the Appendix~\ref{sec:Training_Details_and_Configurations}.

\textbf{Evaluation Benchmarks.}
In simulated environments, we evaluate StereoNav in the Habitat simulator~\cite{Habitat} on Matterport3D~\cite{Matterport3D} scenes, following the Val-Unseen of R2R-CE and RxR-CE. Following standard practice~\cite{NaVid}, we report Navigation Error (NE), Success Rate (SR), Oracle Success Rate (OSR) and Success weighted by Path Length (SPL), which together measure goal attainment, trajectory efficiency, and path fidelity. In real-world environments, we use Success Rate as the primary metric and evaluate StereoNav on a Unitree G1 robot equipped with a ZED Mini stereo camera across four representative navigation scenarios: Office, Gym, Lobby, and Outdoor. A trial is considered successful if the robot stops within 1.5 meters of the goal.

\subsection{Comprehensive Performance Evaluation}
\label{sec:comprehensive_performance_evaluation}

We comprehensively evaluate StereoNav in both simulated and real-world environments. In simulation, we compare it with existing methods on standard VLN-CE benchmarks under established protocols. In the real world, we assess its practical navigation capability across representative physical scenarios. Detailed settings are provided in Appendix~\ref{sec:Details_Comprehensive_Performance_Evaluation}.

% We comprehensively evaluate StereoNav in both simulated and real-world environments. In simulation, we conduct quantitative comparisons with existing methods on standard VLN-CE datasets under established evaluation protocols. In real-world settings, we further evaluate its practical navigation capability across representative physical scenarios. Detailed experimental settings for the simulated and real-world evaluations are provided in Appendix~\ref{sec:Details_Comprehensive_Performance_Evaluation}.

\textbf{Evaluation in Simulated Environments.} Table~\ref{tab:main} compares StereoNav with recent SOTA methods on R2R/RxR-CE Val-Unseen. On R2R-CE, StereoNav achieves the best overall performance across all observation settings with only a lightweight 3B egocentric RGB agent, improving over the strongest prior result by 13.9\% in SR and 3.6\% in SPL, while reducing NE by 1.7m. Under the same egocentric RGB setting, the gains further increase to 16.4\% in SR, 9.8\% in SPL, 8.7\% in OSR, and 2.0m in NE. Notably, even without external data, StereoNav still achieves SOTA performance among egocentric RGB agents on R2R-CE, improving OSR, SR, and SPL by 2.9\%, 8.1\%, and 2.1\%, respectively. On RxR-CE, while StereoNav does not surpass panoramic or RGB-D agents, it achieves the highest SR of 67.5\% within the egocentric RGB setting. These results demonstrate the effectiveness of our visual rendering, stereo-based unified understanding, and joint prediction designs.

% \textbf{Evaluation in Simulated Environments.} Table~\ref{tab:main} compares StereoNav with recent SOTA methods on R2R/RxR-CE Val-Unseen. We report two variants trained with standard navigation data only and with additional external data, and use the latter for direct comparison since prior methods are reported under their default training protocols with additional data. On R2R-CE, StereoNav achieves the best overall performance across all observation settings with only a lightweight 3B egocentric RGB agent, improving over the strongest prior result by 13.9\% in SR and 3.6\% in SPL, while reducing NE by 1.7m. Under the same egocentric RGB setting, the gains further increase to 16.4\% in SR, 9.8\% in SPL, 8.7\% in OSR, and 2.0m in NE. Notably, even without external data, StereoNav still achieves SOTA performance among egocentric RGB agents on R2R-CE, improving OSR, SR, and SPL by 2.9\%, 8.1\%, and 2.1\%, respectively. On RxR-CE, while StereoNav does not surpass panoramic or RGB-D agents, it achieves the highest SR of 67.5\% within the egocentric RGB setting. These results demonstrate the effectiveness of StereoNav for practical egocentric RGB navigation, where robust spatial grounding, stereo-based understanding, and joint prediction are crucial for reliable decision-making.

\begin{figure*}[!t]
    \centering
    \includegraphics[width=\linewidth]{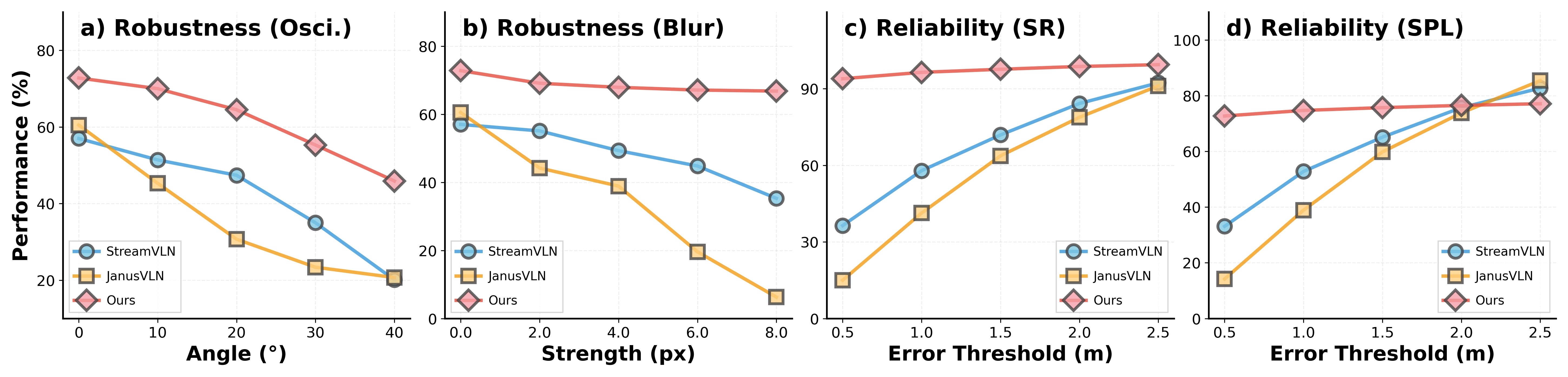}
    \vspace{-0.7cm}
    \caption{\textbf{Robustness and reliability evaluation of StereoNav.}
(a--b) Robustness under viewpoint oscillation and motion blur, where StereoNav shows smaller performance degradation as perturbation severity increases. 
(c--d) Reliability in goal stopping under different stopping-error thresholds, where StereoNav achieves higher SR and SPL, especially under strict target-neighborhood constraints.}
    \vspace{-1.0em}
    \label{fig:robustness_reliability}
\end{figure*}

\textbf{Evaluation in Real-world Environments.} We further evaluate real-world deployment. Each scenario includes three simple and two complex instructions, and each instruction is repeated three times. As shown in Table~\ref{tab:real_world_ablation} (a), StereoNav consistently surpasses both zero-shot VLN methods~\cite{Open-Nav,DreamNav} and supervised egocentric baselines~\cite{NaVid,StreamVLN,JanusVLN}. Macro-averaged over all scenario-difficulty settings, StereoNav achieves a success rate of 60.6\%, markedly higher than the strongest supervised baseline StreamVLN (24.3\%) and the strongest zero-shot baseline DreamNav (22.1\%). More importantly, this advantage persists under complex instructions and more challenging deployment scenarios, such as Lobby and Outdoor, where most baselines degrade to near-zero success. These results demonstrate that StereoNav provides stronger cross-scene generalization and more reliable real-world navigation. Qualitative examples are shown in Fig.~\ref{fig:robustness_reliability}.

\begin{table}[!t]
    \centering
    \vspace{-0.8cm}
    \caption{\textbf{Real-world performance and architectural ablation.}
    (a) \textbf{Left:} Success rates of different methods across four real-world scenarios with simple and complex instructions.
    (b) \textbf{Right:} Ablation of architectural designs in StereoNav, reporting navigation performance under model variants.}
    \label{tab:real_world_ablation}

    \begin{minipage}[t]{0.518\columnwidth}
        \centering
        \scriptsize
        \resizebox{\linewidth}{!}{
            % \begin{tabular}{c|ccccc}
%     \toprule
%     \rowcolor{gray!15}
%     & \multicolumn{5}{c}{\textbf{Scenario}} \\
%     \rowcolor{gray!15}
%     \multirow{-2}{*}{\cellcolor{gray!15}\textbf{Method}} 
%     & \textbf{~~Office~~} & \textbf{~~Gym~~} & \textbf{~~Lobby~~} & \textbf{~~Outdoor~~} & \textbf{~~Average~~} \\
%     \midrule
%     \midrule
%     Open-Nav~\cite{Open-Nav} & - & - & - & - & - \\
%     DreamNav~\cite{DreamNav} & - & - & - & - & - \\
%     \midrule
%     NaVid~\cite{NaVid} & - & - & - & - & - \\
%     StreamVLN~\cite{StreamVLN} & - & - & - & - & - \\
%     JanusVLN~\cite{JanusVLN}  & - & - & - & - & - \\
%     \midrule
%     StereoNav & - & - & - & - & - \\
%     \bottomrule
% \end{tabular}

% 每个场景 3 个简单指令 2 个难指令
% 每个重复 3次

\begin{tabular}{c|cc|cc|cc|cc}
    \toprule
    \rowcolor{gray!15}
    & \multicolumn{2}{c|}{\textbf{Office}} 
    & \multicolumn{2}{c|}{\textbf{Gym}} 
    & \multicolumn{2}{c|}{\textbf{Lobby}} 
    & \multicolumn{2}{c}{\textbf{Outdoor}} \\
    \rowcolor{gray!15}
    \multirow{-2}{*}{\cellcolor{gray!15}\textbf{Method}} 
    & \textbf{Sim.} & \textbf{Com.}
    & \textbf{Sim.} & \textbf{Com.}
    & \textbf{Sim.} & \textbf{Com.}
    & \textbf{Sim.} & \textbf{Com.} \\
    \midrule
    \midrule
    Open-Nav~\cite{Open-Nav} 
    & 0.33 & 0.00 & 0.22 & 0.00 & 0.22 & 0.00 & 0.33 & \underline{0.17} \\
    DreamNav~\cite{DreamNav} 
    & 0.44 & 0.00 & 0.33 & 0.17 & 0.22 & 0.00 & \underline{0.44} & \underline{0.17} \\
    \midrule
    NaVid~\cite{NaVid} 
    & 0.33 & 0.00 & 0.33 & 0.00 & 0.11 & 0.00 & 0.00 & 0.00 \\
    StreamVLN~\cite{StreamVLN} 
    & \underline{0.56} & \underline{0.17} & \underline{0.44} & \underline{0.33} & \underline{0.33} & 0.00 & 0.11 & 0.00 \\
    JanusVLN~\cite{JanusVLN}  
    & 0.22 & 0.00 & 0.11 & 0.17 & \underline{0.33} & 0.00 & 0.22 & 0.00 \\
    \midrule
    StereoNav 
    & \textbf{0.67} & \textbf{0.50} & \textbf{0.56} & \textbf{0.83} & \textbf{0.78} & \textbf{0.17} & \textbf{0.67} & \textbf{0.67} \\
    \bottomrule
\end{tabular}
        }
    \end{minipage}
    \hfill
    \begin{minipage}[t]{0.475\columnwidth}
        \centering
        \scriptsize
        \resizebox{\linewidth}{!}{

% \begin{tabular}{ccc|cc|cccc}
%     \toprule
%     \rowcolor{gray!15}
%     \multicolumn{3}{c|}{\textbf{Architecture}} & 
%     \multicolumn{2}{c|}{\textbf{Views}} & 
%     \multicolumn{4}{c}{\textbf{R2R-CE Val-Unseen}} \\
%     \rowcolor{gray!15}
%     \textbf{3D Geo.} & \textbf{2D Stru.} & \textbf{2D Sema.} & 
%     \textbf{Stereo} & \textbf{Rend.} &
%     \textbf{NE$\downarrow$} & \textbf{OSR$\uparrow$} & \textbf{SR$\uparrow$} & \textbf{SPL$\uparrow$} \\
%     \midrule\midrule
%      &  & \ding{51} & L/- & \ding{55} & 8.7 & 42.0 & 24.6 & 21.1 \\
%      &  & \ding{51} & L/- & \ding{51} & 4.1 & 61.0 & 58.1 & 51.9 \\
%      \midrule
%       & \ding{51} & \ding{51} & R/R & \ding{51} & - & - & - & - \\
%       & \ding{51} & \ding{51} & L/R & \ding{51} & - & - & - & - \\
%      \ding{51} &  & \ding{51} & L/R & \ding{51} & 4.4 & 54.5 & 51.3 & 46.0 \\
%      \midrule
%     \ding{51} & \ding{51} & \ding{51} & L/R & \ding{51} & 3.0 & 76.6 & 72.8 & 56.4 \\
%     \bottomrule
% \end{tabular}

\begin{tabular}{cccc|cc|cc}
    \toprule
    \rowcolor{gray!15}
    \multicolumn{4}{c|}{\textbf{Architecture}} & 
    \multicolumn{2}{c|}{\textbf{Views}} & 
    \multicolumn{2}{c}{\textbf{R2R-CE}} \\
    \rowcolor{gray!15}
    \textbf{3D Geo.} & \textbf{2D Stru.} & \textbf{2D Sema.} & \textbf{De. Head} & 
    \textbf{Stereo} & \textbf{Rend.} &
    \textbf{SR$\uparrow$} & \textbf{SPL$\uparrow$} \\
    \midrule\midrule
    \ding{55} & \ding{55} & \ding{51} & \ding{55} & L/- & \ding{55} & 34.4 & 29.5 \\
    \ding{55} & \ding{55} & \ding{51} & \ding{55} & L/- & \ding{51} & 52.3 & 46.7 \\
    \midrule
    \ding{55} & \ding{51} & \ding{51} & \ding{55} & R/R & \ding{51} & 48.3 & 37.0 \\
    \ding{55} & \ding{51} & \ding{51} & \ding{55} & L/R & \ding{51} & \underline{63.4} & \underline{51.7} \\
    \ding{51} & \ding{55} & \ding{51} & \ding{55} & L/R & \ding{51} & 51.3 & 46.0 \\
    \ding{51} & \ding{55} & \ding{51} & \ding{51} & L/R & \ding{51} & 58.4 & 50.6 \\
    \midrule
    \ding{51} & \ding{51} & \ding{51} & \ding{51} & L/R & \ding{51} & \textbf{72.8} & \textbf{56.4} \\
    \bottomrule
\end{tabular}
        }
    \end{minipage}

\end{table}

\subsection{Robustness and Reliability Evaluation}
\label{sec:robustness_reliability_evaluation}
We further assess StereoNav under deployment-oriented conditions, evaluating its robustness to severe visual disturbances and its ability to terminate accurately near the target. Additional experiments and detailed settings are provided in Appendix~\ref{sec:Details_Robustness_Reliability_Evaluation}.

As shown in Fig.~\ref{fig:robustness_reliability}, StereoNav consistently outperforms prior VLN agents in robustness and reliability. Under the strongest viewpoint oscillation, it incurs a 37.0\% relative degradation, much lower than StreamVLN (64.7\%) and JanusVLN (65.8\%); under motion blur, the degradation is only 8.2\%, compared with 38.1\% and 89.6\%. These results show that stereo-based unified understanding and joint action-depth prediction improve perceptual stability under corrupted observations. StereoNav also yields more reliable goal stopping: under a strict 0.5\,m threshold, 93.9\% of successful episodes remain within the target neighborhood, with 72.7\% SPL, substantially surpassing StreamVLN (36.4\%/33.1\%) and JanusVLN (15.0\%/14.2\%) in SR/SPL. This confirms that the rendered target prior provides persistent visual guidance for resolving ambiguous routes and stopping decisions.

% As shown in Fig.~\ref{fig:robustness_reliability}, StereoNav consistently exhibits stronger robustness and reliability than prior VLN agents. Under the strongest viewpoint oscillation, StereoNav shows a 37.0\% relative degradation, substantially lower than StreamVLN (64.7\%) and JanusVLN (65.8\%). This advantage is even more pronounced under motion blur, where StereoNav degrades by only 8.2\%, compared with 38.1\% and 89.6\% for StreamVLN and JanusVLN. These results indicate that stereo-based unified understanding and joint action-depth prediction improve perceptual stability under corrupted observations. StereoNav also achieves more reliable goal stopping: under a strict 0.5\,m threshold, 93.9\% of successful episodes remain within the target neighborhood, with 72.7\% SPL, far surpassing StreamVLN (36.4\%/33.1\%) and JanusVLN (15.0\%/14.2\%) in SR/SPL. This confirms that the rendered target prior provides effective persistent visual guidance for resolving ambiguous route and stopping decisions.

\subsection{Ablation Study}
\label{sec:ablation_study}
We conduct ablation studies on a subset of the R2R-CE evaluation split to analyze the contribution of each key component in StereoNav. More detailed ablation results are provided in Appendix~\ref{sec:Additional_Ablation_Studies}.

\textbf{Ablation Study on Different Model Designs.} Table~\ref{tab:real_world_ablation} (b) shows that the performance gains of StereoNav arise from the coordinated design of visual guidance, Unified Understanding Modeling, and Joint Prediction Modeling. The rendered target-point prior improves SR/SPL by 17.9\%/17.2\% over the monocular semantic baseline, confirming that persistent visual guidance helps mitigate instruction ambiguity. With this guidance, Unified Understanding Modeling further improves navigation by integrating semantic, structural, and geometric cues, and the full model achieves the best performance of 72.8\% SR and 56.4\% SPL. Further comparisons show that stereo-based 2D modeling contributes 15.1\%/14.7\% gains in SR/SPL, while depth-supervised geometry adds another 7.1\%/4.6\%, indicating that binocular cues and joint action-depth prediction improve navigation-relevant spatial reasoning.

% \textbf{Ablation Study on the Effect of Different Model Designs.} As shown in Table~\ref{tab:real_world_ablation} (b), each design in StereoNav contributes to the final performance. Introducing the rendered target-point prior substantially improves the monocular semantic baseline, raising SR/SPL from 34.4/29.5 to 52.3/46.7, which shows that persistent visual guidance can effectively compensate for instruction ambiguity. Beyond this guidance design, the results further validate the importance of Unified Understanding Modeling. Compared with semantic-only or partially unified variants, the full model that integrates semantic, structural, and geometric cues achieves the best performance, reaching 72.8 SR and 56.4 SPL, indicating that these cues provide complementary information for navigation. Within this unified design, stereo-based 2D modeling is also critical: applying the semantic and structural encoders to different stereo views improves SR/SPL from 48.3/37.0 to 63.4/51.7, suggesting that binocular view encoding implicitly preserves useful disparity cues and enhances spatial discrimination. Moreover, while directly adding 3D geometric tokens brings limited benefit, combining them with depth supervision improves SR/SPL from 51.3/46.0 to 58.4/50.6, showing that joint action-depth prediction helps make geometric representations more navigation-relevant. Overall, these ablations indicate that the proposed designs improve navigation performance in a complementary manner.

\section{Conclusion}
In this work, we revisit the limitations of Vision-and-Language Navigation beyond the conventional focus on stronger model understanding, identifying visual uncertainty and instructional under-specification as two key bottlenecks. We introduce StereoNav, a stereo Vision-Language-Action framework that renders target-location priors as persistent visual guidance, unifies semantic, structural, and geometric cues for robust spatial understanding, and couples action generation with depth prediction for geometry-aware decision making. Evaluations across simulation, real-world deployment, robustness, and reliability demonstrate that StereoNav achieves state-of-the-art performance. However, StereoNav currently relies on explicit or implicit target-oriented input and cannot yet handle cases where such information is absent. Future work will explore multimodal grounding and spatial estimation to infer target-location priors directly from observations and language.

\bibliography{reference}
\bibliographystyle{unsrtnat}

\clearpage
\newpage
\appendix

\section{Details of the Pilot Studies}

\subsection{Experimental Setup for Visual Uncertainty}
\label{sec:Visual_Uncertainty_Experimental_Setup}

To examine whether the performance degradation of current VLN agents is associated with visual uncertainty, we conduct controlled pilot studies on two representative open-source VLM-based navigation methods. Specifically, we choose StreamVLN~\cite{StreamVLN} as the LLaVA-based method~\cite{LLaVA_Video}, which adopts LLaVA-NeXT as its backbone, and JanusVLN~\cite{JanusVLN} as the Qwen-based method, which is built upon Qwen2.5-VL~\cite{Qwen2_5_VL}. These two methods represent recent VLN agents based on widely used multimodal backbones, allowing us to evaluate whether visual disturbances consistently affect different VLM-based navigation pipelines. To isolate the effect of visual uncertainty, we keep the navigation instructions, evaluation episodes, action space, and evaluation metrics unchanged, and only modify the visual observations received by the agent during inference.

We manually reproduce four common forms of visual uncertainty in embodied navigation: lighting perturbation, height fluctuation, motion blur, and viewpoint oscillation. Lighting perturbation is implemented by post-processing the RGB observations with a brightness scaling factor of $-0.8$, which darkens the image by approximately $80\%$ and simulates under-exposed visual inputs. Height fluctuation is implemented by increasing the height of the RGB camera sensors by $0.6$ m, thereby changing the egocentric viewpoint while keeping the navigation task unchanged. Motion blur is applied before image resizing with a blur strength of $8.0$, simulating image degradation caused by robot motion or camera shake. Viewpoint oscillation is implemented by temporarily perturbing the camera orientation in the simulator and alternately rotating the egocentric view to the left and right by $40^\circ$ at consecutive steps. These perturbation levels are intentionally set as challenging stress-test conditions, so that the resulting performance changes can directly reflect the robustness of existing VLN agents under deployment-oriented visual disturbances.

\subsection{Experimental Setup for Instruction Ambiguity}
\label{sec:Instruction_Ambiguity_Experimental_Setup}

\begin{figure*}[!b]
    \centering 
    \vspace{-1.0em}
    \includegraphics[width=\linewidth]{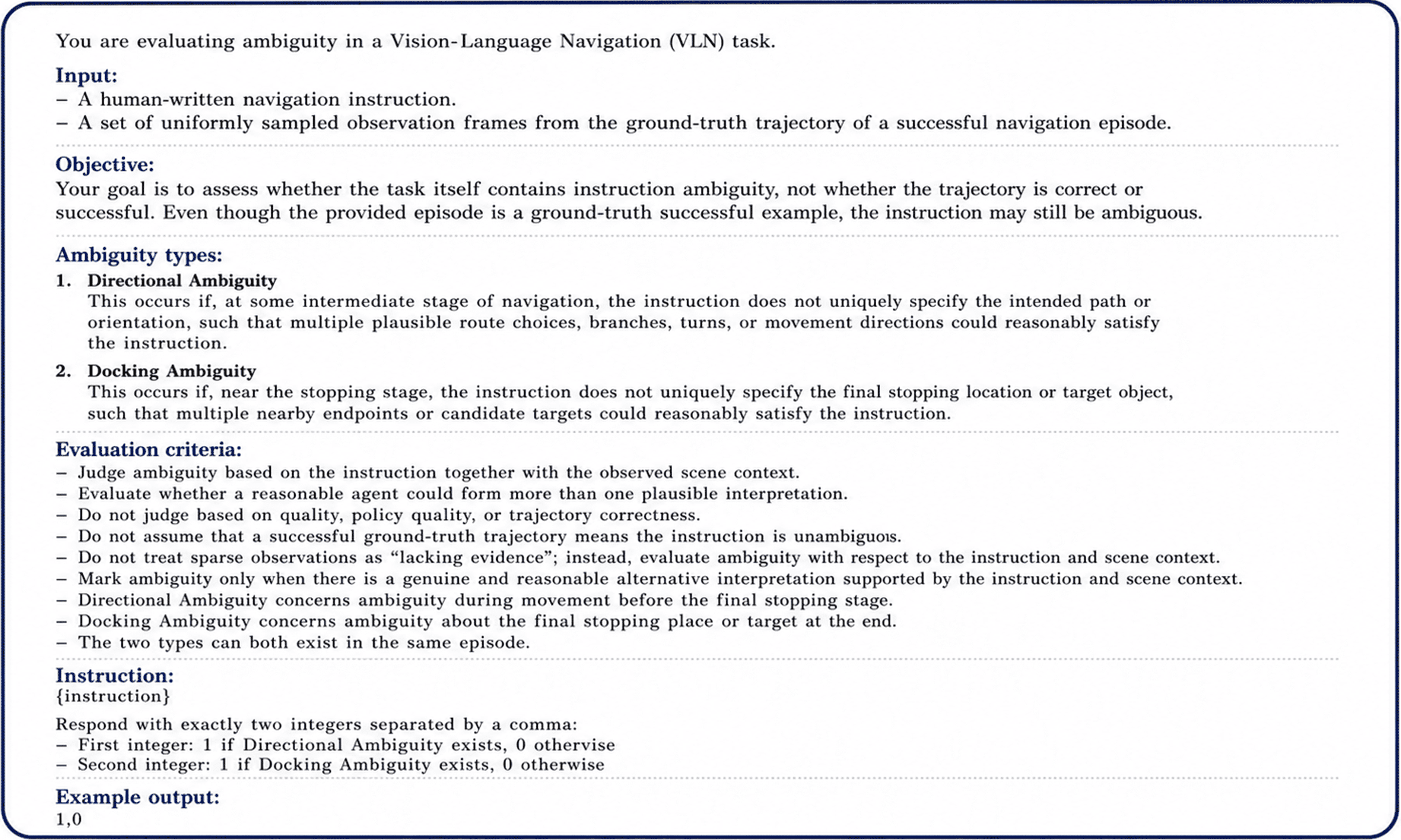}
    \caption{\textbf{Prompt template for instruction ambiguity assessment.} Given a navigation instruction and sampled observation frames, the evaluator identifies whether Directional Ambiguity or Docking Ambiguity exists and returns binary labels for the two ambiguity types.}
    \vspace{-1.0em}
    \label{fig:appendix_1}
\end{figure*}

To analyze the relationship between instruction ambiguity and navigation performance, we construct a diagnostic set from the R2R-CE Val-Unseen split using StreamVLN~\cite{StreamVLN} rollouts. Specifically, we randomly sample 100 successful episodes and 100 failed episodes. For each episode, we extract the original navigation instruction and the visual observations collected along the expert trajectory, so that ambiguity can be assessed with respect to the intended route rather than the agent's executed trajectory. To reduce temporal redundancy while preserving the main route context, we uniformly sample 10 frames from each observation sequence and pair them with the corresponding instruction. The sampled frames and instruction are then organized into a unified evaluation prompt, as shown in Fig.~\ref{fig:appendix_1}, where the evaluator is asked to determine whether the instruction is affected by Directional Ambiguity or Docking Ambiguity.

We perform the ambiguity assessment with three strong multimodal models, including GPT-5~\cite{GPT-5}, Gemini-2.5-Pro~\cite{Gemini_2_5}, and Claude-Opus-4.7~\cite{Claude}. Each model independently judges the presence of the two ambiguity types for every sampled episode. For each ambiguity type and each outcome group, we first compute the percentage of episodes judged as being affected by that ambiguity, denoted as $r$. We then define the ambiguity score as $100-r$, where a lower score indicates that a larger proportion of episodes is affected by the corresponding ambiguity type. The final results are reported for each evaluator and further averaged across the three models, providing a more robust estimate of how directional and docking ambiguity correlate with navigation success and failure.

\section{Details of the StereoNav}
\subsection{Visual Rendering}

\label{sec:Appendix_Visual_Rendering}
We provide the implementation details of target-prior rendering in Algorithm~\ref{alg:target_prior_rendering}. At each time step, StereoNav first transforms the target prior from the world frame to each stereo camera frame using the current agent pose and the corresponding camera offset. The target is then projected onto the image plane with the camera intrinsics. We define a valid image region and a relaxed boundary region to handle near-boundary projections. If the projected point is valid and falls within the relaxed region, it is clipped to the image range and rendered as the target cue. Otherwise, the cue is placed at the center of the left or right image boundary according to the horizontal direction of the target in the camera frame. This procedure is applied to both stereo views, producing target-aware observations that are consistent across data generation and evaluation.

\begin{algorithm}[htbp]
\caption{Visual Rendering}
\label{alg:target_prior_rendering}
\small
\DontPrintSemicolon
\SetAlgoNoLine
\SetKwInOut{Input}{Input}
\SetKwInOut{Output}{Output}
\SetKwInOut{Notation}{Notation}

\Input{
Stereo observations $(o_t^l,o_t^r)$, target-location prior $p$, agent pose 
$\mathcal{P}_t=(\mathbf{x}_t,\mathbf{q}_t)$, camera intrinsics $K$, image size $(W,H)$, 
stereo camera offsets $(\Delta^l,\Delta^r)$, and relaxation factor $\epsilon$.
}

\BlankLine

\Output{
Rendered stereo observations $(\tilde{o}_t^l,\tilde{o}_t^r)$.
}

\BlankLine

\textbf{Define 1:} $\Omega=[0,W-1]\times[0,H-1]$, the valid image domain.\;
\textbf{Define 2:} $\Omega_{\epsilon}=[-\epsilon W,(1+\epsilon)W-1]\times[-\epsilon H,(1+\epsilon)H-1]$, the relaxed projection domain.\;

\BlankLine

\ForEach{view $v\in\{l,r\}$}{
    Compute the camera center:
    $
    \mathbf{c}_t^v=\mathbf{x}_t+R(\mathbf{q}_t)\Delta^v
    $\;

    Transform the target prior into the camera coordinate frame:
    $
    \mathbf{s}_t^v=(X_t^v,Y_t^v,Z_t^v)
    =R(\mathbf{q}_t)^\top(p-\mathbf{c}_t^v)
    $\;

    \uIf{$Z_t^v>0$}{
        Project the $p$ onto the image plane:
        $
        \boldsymbol{\pi}_t^v=\Pi_K(\mathbf{s}_t^v)
        $\;
    }
    \Else{
        Set $\boldsymbol{\pi}_t^v=\varnothing$\;
    }
    
    \uIf{$\boldsymbol{\pi}_t^v=\varnothing$ and $\boldsymbol{\pi}_t^v\in\Omega_\epsilon$}{
        Clip the projection to the valid image domain:
        $
        \boldsymbol{\rho}_t^v \leftarrow \operatorname{clip}(\boldsymbol{\pi}_t^v,\Omega)
        $\;
    }
    \Else{
        \eIf{$X_t^v<0$}{
            $\boldsymbol{\rho}_t^v \leftarrow (0,H/2)$\;
        }{
            $\boldsymbol{\rho}_t^v \leftarrow (W-1,H/2)$\;
        }
    }

    Render a red circular marker centered at $\boldsymbol{\rho}_t^v$ on $o_t^v$ to obtain $\tilde{o}_t^v$\;
}

\Return{$(\tilde{o}_t^l,\tilde{o}_t^r)$}\;
\end{algorithm}

\subsection{Prompt Template}
\label{sec:Prompt_Template}
\begin{figure*}[!t]
    \centering 
    \vspace{-1.0em}
    \includegraphics[width=\linewidth]{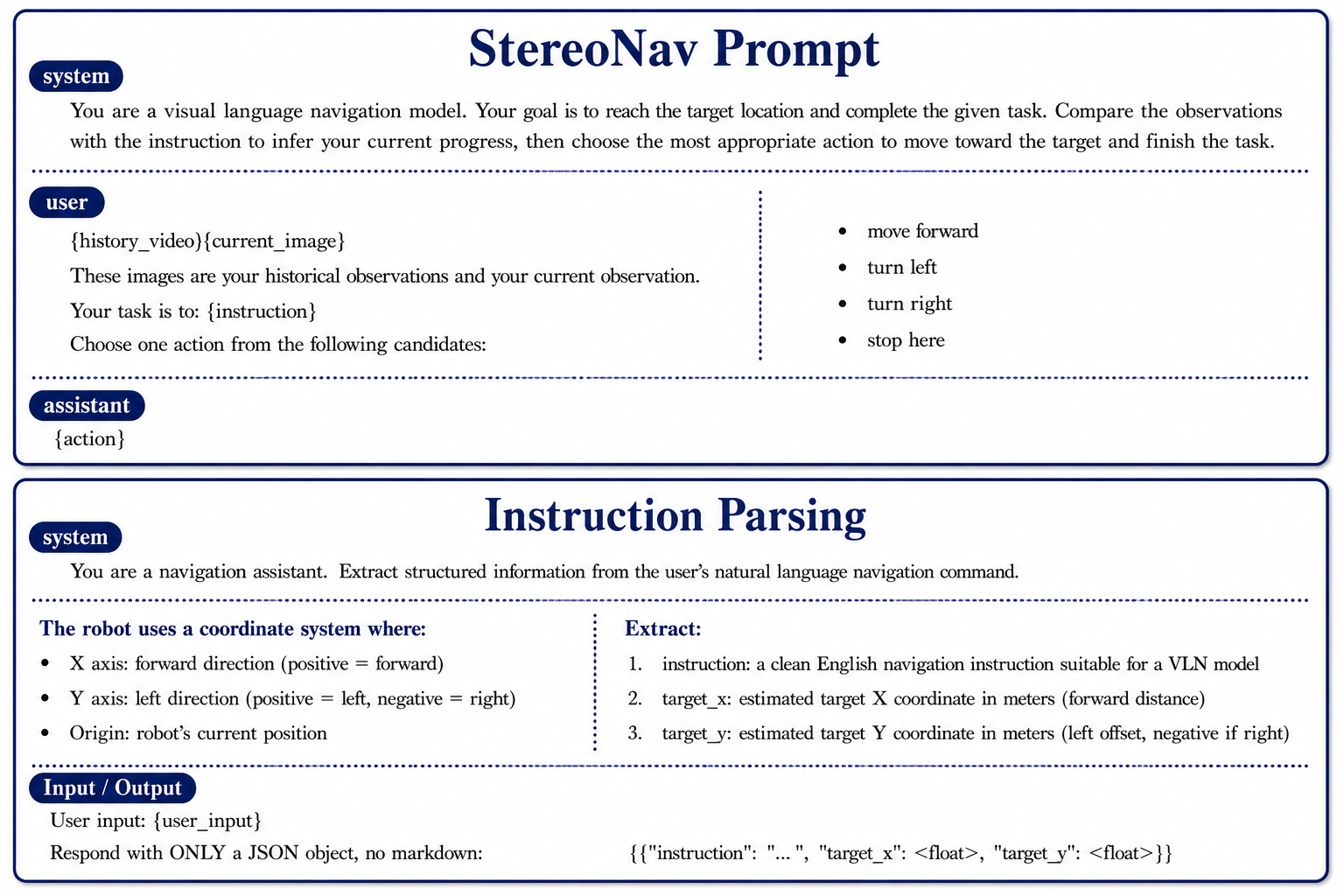}
    \caption{\textbf{Prompt templates for StereoNav model input and real-world instruction preprocessing.}
(a) \textbf{Top:} The StereoNav model prompt formats observations and instructions as inputs, with the assistant response as the action label.
(b) \textbf{Bottom:} The real-world preprocessing prompt converts user commands into the structured instruction and target-location prior required by StereoNav.}
    \vspace{-1.0em}
    \label{fig:appendix_2}
\end{figure*}

We provide the detailed prompt template used by StereoNav in Fig.~\ref{fig:appendix_2}(a). The prompt follows a conversation-style format and consists of three parts: a system message that defines the navigation role and action-selection objective, a user message that provides historical observations, the current observation, and the navigation instruction, and an assistant response that corresponds to the target action. In this formulation, visual inputs are organized as historical frames and the current frame, while the language instruction specifies the navigation goal. The model is required to select one action from a fixed action space, including moving forward, turning left, turning right, and stopping. This prompt design keeps the navigation interface simple and consistent across training and inference, allowing StereoNav to learn action prediction from multimodal context in a unified textual format.

\subsection{Training Details and Configurations}
\label{sec:Training_Details_and_Configurations}
\begin{table*}[t]
    \centering
    \small
    \caption{\textbf{Training configurations of StereoNav.} The table summarizes the main settings used in the two-stage training procedure, including training setup, model configuration, and optimization details.}
    \label{tab:training_config}
    \vspace{1em}
    \resizebox{\linewidth}{!}{
    \begin{tabular}{c|cc}
        \toprule
        \textbf{~~~~~~~~~~~~~~~~Configuration~~~~~~~~~~~~~~~~} & \textbf{~~~~~~~~~~~~~~~~~~~~~~~~~~~~~Stage I~~~~~~~~~~~~~~~~~~~~~~~~~~~~~} & \textbf{~~~~~~~~~~~~~~~~~~~~~~~~~Stage II~~~~~~~~~~~~~~~~~~~~~~~~~} \\
        \midrule\midrule

        \rowcolor{gray!15}\multicolumn{3}{c}{\textit{Basic Training Setup}} \\

        Precision 
        & bfloat16 
        & bfloat16 \\

        Global Batch Size 
        & 512 
        & 512 \\

        Training Schedule 
        & 1 epoch 
        & 1 epoch \\

        Training Data 
        & R2R, RxR 
        & R2R, RxR, DAgger, ScaleVLN \\

        Sliding Window 
        & 24 frames, $\beta{=}8$ 
        & 24 frames, $\beta{=}6$ \\

        Training Cost
        & ~176 GPU hours
        & ~688 GPU hours \\
        
        \midrule
        \rowcolor{gray!15}\multicolumn{3}{c}{\textit{Model Setup}} \\
        Input Resolution 
        & $448{\times}448$ 
        & $448{\times}448$ \\

        History Length 
        & 8 frames 
        & 8 frames \\

        Prediction Horizon 
        & 4 actions 
        & 4 actions \\

        Maximum Token Length 
        & 4096 
        & 4096 \\

        Depth Configuration
        & 5 m range, 4 iterations
        & 5 m range, 4 iterations \\

        Fusion Weights 
        & 1.0, 0.6, 0.2 
        & 1.0, 0.6, 0.2 \\

        Initialization 
        & InternVL3.5, DINOv2, FoundationStereo 
        & Stage-I checkpoint \\

        \midrule
        \rowcolor{gray!15}\multicolumn{3}{c}{\textit{Optimization and Loss Setup}} \\
        Optimizer 
        & AdamW 
        & AdamW \\

        Gradient Clipping 
        & 0.5 
        & 0.5 \\

        Base Learning Rate 
        & $2.0{\times}10^{-5}$ 
        & $2.0{\times}10^{-5}$ \\

        Depth Learning Rate 
        & $2.0{\times}10^{-6}$ 
        & $2.0{\times}10^{-6}$ \\

        Scheduler 
        & warm up 0.05, min lr 0.9 
        & warm up 0.05, min lr 0.5 \\

        Loss Weights 
        & $\lambda_{\mathrm{lang}}{=}1.0$, $\lambda_{\mathrm{depth}}{=}0.1$ 
        & $\lambda_{\mathrm{lang}}{=}1.0$, $\lambda_{\mathrm{depth}}{=}0.1$ \\
        \bottomrule
    \end{tabular}}
\end{table*}

To facilitate reproducibility, we provide the detailed training configurations of StereoNav in Table~\ref{tab:training_config}. The training procedure follows a two-stage design. Stage I focuses on establishing the basic target-aware stereo navigation capability using standard VLN-CE training data, where the model learns to associate rendered target priors, stereo observations, and language instructions with navigation actions. Stage II further initializes from the Stage-I checkpoint and incorporates additional trajectory data to improve data coverage and policy robustness. The table summarizes the key implementation choices across the two stages, including the basic training setup, model input and prediction settings, depth-related configuration, and optimization objectives. These details are intended to clarify the training protocol without overloading the main paper with implementation-specific hyperparameters.

\section{Details of Comprehensive Performance Evaluation}
\label{sec:Details_Comprehensive_Performance_Evaluation}

\subsection{Simulation Evaluation}
\label{sec:simulation_evaluation}

\begin{figure*}[htbp]
    \centering 
    \includegraphics[width=\linewidth]{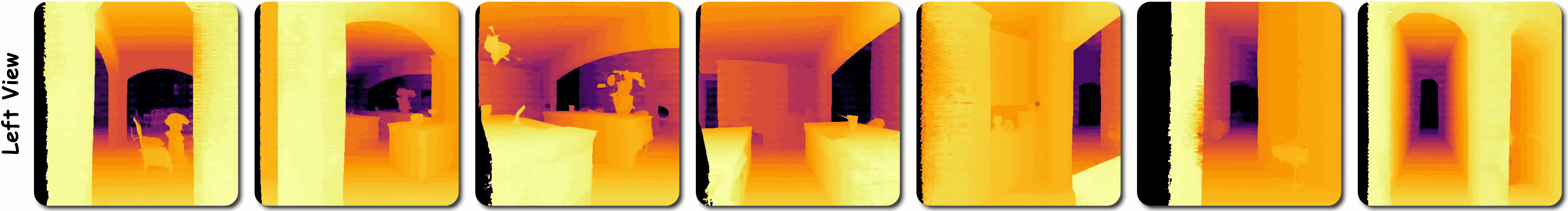}
    \caption{\textbf{Visualization of depth estimation results in a representative virtual environment.} StereoNav leverages stereo disparity between the left and right first-person views to produce stable depth estimates, providing reliable geometric cues for robust navigation under approximate goal priors.}
    \label{fig:appendix_4}
\end{figure*}

\begin{figure*}[htbp]
    \centering 
    \includegraphics[width=\linewidth]{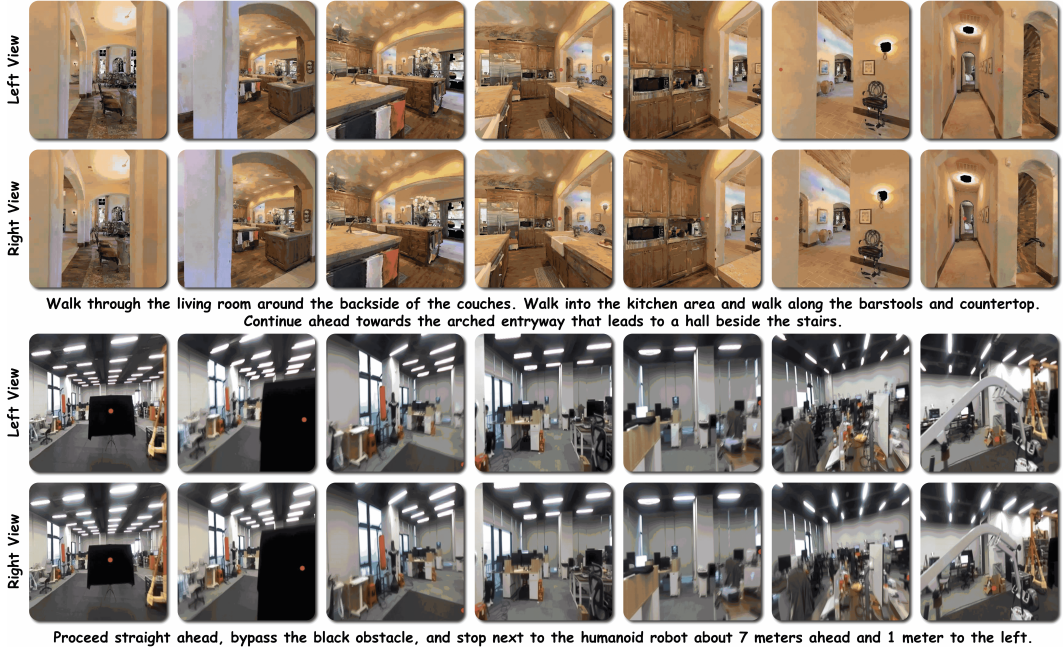}
    \caption{\textbf{First-person stereo observations in simulation and real-world deployment.}
    (a) \textbf{Top:} First-person left- and right-view observations from a representative virtual environment, where the rendered target-location prior provides persistent visual guidance during navigation.
    (b) \textbf{Bottom:} First-person left- and right-view observations from a representative real-world Gym scenario, demonstrating the deployment of StereoNav on the Unitree G1 robot under physical navigation conditions.}
    \label{fig:appendix_3}
\end{figure*}

We conduct all simulation experiments using Habitat-Lab with Habitat-Sim as the underlying simulator, following the standard VLN-CE evaluation protocol. The agent operates in continuous 3D environments while executing a discrete action space, including moving forward, turning left, turning right, and stopping. In the simulator, the agent is represented as a cylindrical collision body with a height of 1.5 m and a radius of 0.1 m. Stereo RGB observations are captured by two pinhole cameras mounted at a height of 1.25 m, with a stereo baseline of 0.1 m. During inference, we adopt deterministic decoding with \texttt{temperature}=0.0 and \texttt{top\_p}=1.0. Each episode terminates when the agent predicts the stop action or reaches the maximum episode length of 500 steps. At the top of Fig.~\ref{fig:appendix_3}, we further present first-person views from a representative virtual environment. We also visualize the corresponding depth estimation results in Fig.~\ref{fig:appendix_4}. Under this unified simulation protocol, we further evaluate StereoNav with controlled target-location deviations to analyze its robustness under fuzzy goal priors.

\begin{table}[htbp]
    \centering
    \caption{\textbf{Performance of StereoNav under controlled target-location deviations on the R2R-CE Val-Unseen split.} The preset value denotes the maximum radius used to randomly sample a perturbed target point around the ground-truth goal, while the actual value reports the average Euclidean deviation of the sampled targets. The first row corresponds to the clean setting, and values in parentheses indicate the absolute performance difference from this setting.}
    \label{tab:appendix_coarse}
    \small
    \setlength{\tabcolsep}{6pt}
    \renewcommand{\arraystretch}{1.12}
    \resizebox{0.75\columnwidth}{!}{
    \begin{tabular}{c|cc|cccc}
        \toprule
        \rowcolor{gray!15}
        \cellcolor{gray!15}
        & \multicolumn{2}{c|}{\textbf{Deviation}} 
        & \multicolumn{4}{c}{\textbf{R2R-CE Val-Unseen}} \\
        \rowcolor{gray!15}
        \multirow{-2}{*}{\cellcolor{gray!15}\textbf{Method}}
        & \textbf{Preset}
        & \textbf{Actual}
        & \textbf{~~NE$\downarrow$~~}
        & \textbf{~~OSR$\uparrow$~~} 
        & \textbf{~~SR$\uparrow$~~} 
        & \textbf{~~SPL$\uparrow$~~} \\
        \midrule\midrule
        \rowcolor{blue!4}
        \multirow{4}{*}{StereoNav}
        & 0.0 & 0.0
        & 3.0 
        & 76.6 
        & 72.8 
        & 56.4 \\
        & 1.0 & 0.7 
        & 3.2{\scriptsize\,\textcolor{red!70!black}{($\uparrow$0.2)}} 
        & 74.7{\scriptsize\,\textcolor{red!70!black}{($\downarrow$1.9)}} 
        & 70.3{\scriptsize\,\textcolor{red!70!black}{($\downarrow$2.5)}} 
        & 50.7{\scriptsize\,\textcolor{red!70!black}{($\downarrow$5.7)}} \\
        & 2.0 & 1.4 
        & 4.0{\scriptsize\,\textcolor{red!70!black}{($\uparrow$1.0)}} 
        & 72.4{\scriptsize\,\textcolor{red!70!black}{($\downarrow$4.2)}} 
        & 65.4{\scriptsize\,\textcolor{red!70!black}{($\downarrow$7.4)}} 
        & 42.2{\scriptsize\,\textcolor{red!70!black}{($\downarrow$14.2)}} \\
        & 3.0 & 2.0 
        & 4.9{\scriptsize\,\textcolor{red!70!black}{($\uparrow$1.9)}} 
        & 71.5{\scriptsize\,\textcolor{red!70!black}{($\downarrow$5.1)}} 
        & 50.6{\scriptsize\,\textcolor{red!70!black}{($\downarrow$22.2)}} 
        & 33.0{\scriptsize\,\textcolor{red!70!black}{($\downarrow$23.4)}} \\
        \bottomrule
    \end{tabular}
    }
\end{table}

To evaluate StereoNav under fuzzy target-location priors, we conduct controlled target-location deviation experiments on the R2R-CE Val-Unseen split. Given a preset deviation radius, we randomly sample a perturbed target point within the circle centered at the ground-truth goal. The ``Preset'' column denotes the maximum sampling radius for generating the noisy prior, while ``Actual'' reports the average Euclidean deviation of the sampled targets from the ground-truth location. The first row represents the clean setting, where the provided target point is perfectly aligned with the true goal. As shown in Table~\ref{tab:appendix_coarse}, StereoNav remains effective as the target-location prior becomes increasingly imprecise. In the clean setting, it achieves 3.0 NE, 76.6 OSR, 72.8 SR, and 56.4 SPL. With a 1.0 m preset deviation and a 0.7 m actual deviation, the model shows only mild degradation: NE increases by 0.2 m, while OSR, SR, and SPL decrease by 1.9, 2.5, and 5.7 points, respectively. When the preset deviation increases to 2.0 m, corresponding to a 1.4 m actual deviation, the drops become larger but remain moderate, with OSR, SR, and SPL decreasing by 4.2, 7.4, and 14.2 points. Under the most challenging 3.0 m preset deviation, where the actual deviation reaches 2.0 m, StereoNav still achieves 71.5 OSR, 50.6 SR, and 33.0 SPL. Notably, although SR and SPL drop by 22.2 and 23.4 points, OSR decreases by only 5.1 points, indicating that the agent can still approach the target region in many episodes despite substantial prior noise. These results suggest that the target-location prior serves as complementary spatial guidance rather than an exact coordinate constraint, allowing StereoNav to integrate coarse goal information with egocentric observations, language instructions, and stereo-based spatial understanding for robust navigation under approximate goal cues.

\subsection{Real-World Evaluation}
\label{sec:real_world_evaluation}

We conduct real-world experiments on a Unitree G1 humanoid robot equipped with a ZED Mini stereo camera. The stereo camera is rigidly mounted using the open-source TWIST2 neck firmware and head-mounted structure(\url{https://yanjieze.com/TWIST2/}), which provides stable egocentric stereo observations during navigation. StereoNav is deployed with a cloud-inference architecture: the G1 robot is directly connected to a local laptop via Ethernet, and the laptop bridges communication between the robot platform and the remote inference server. At each step, the laptop streams stereo RGB observations, the navigation instruction, and the rendered target prior to the server, and then sends the predicted high-level action back to the robot for execution through the G1 control interface. To support natural human-robot interaction, free-form user commands are first processed by a GPT-based instruction parser, as shown in Fig.~\ref{fig:appendix_2} (b). The parser converts a raw user command into a clean VLN-style instruction and extracts the corresponding target-location prior required by StereoNav. We further show first-person views from a representative Gym scenario at the bottom of Fig.~\ref{fig:appendix_3}.

\section{Details of Robustness and Reliability Evaluation}
\label{sec:Details_Robustness_Reliability_Evaluation}

In the main paper, we primarily report robustness results under severe visual disturbances, including motion blur and viewpoint oscillation, as these conditions more directly reflect challenging deployment scenarios that can induce substantial perception instability. Here, we provide complementary results under milder but commonly encountered conditions, including lighting perturbations and camera-height fluctuations. For a conservative comparison, we evaluate the StereoNav model trained only on R2R-CE and RxR-CE, and compare it with StreamVLN and JanusVLN using their full models trained with diverse navigation data. This setting allows us to examine whether the robustness advantage of StereoNav also holds under less severe yet practically relevant observation changes.

\begin{table}[htbp]
    \centering
    \caption{\textbf{Robustness under lighting perturbations and camera-height fluctuations on the R2R-CE Val-Unseen split.} We report Success Rate (SR, \%) under lighting perturbations ($-0.8/+0.8$) and camera-height fluctuations ($-0.6/+0.6$ m). Values in parentheses denote the absolute SR drop relative to each method's clean setting.}
    \label{tab:robustness_perturbation_fluctuation}
    \small
    \setlength{\tabcolsep}{6pt}
    \renewcommand{\arraystretch}{1.12}
    \resizebox{0.70\columnwidth}{!}{
    \begin{tabular}{c|cc|cc}
        \toprule
        \rowcolor{gray!15}
        \cellcolor{gray!15}
        & \multicolumn{2}{c|}{\textbf{Perturbation}} 
        & \multicolumn{2}{c}{\textbf{Fluctuation}} \\
        \rowcolor{gray!15}
        \multirow{-2}{*}{\cellcolor{gray!15}\textbf{Method}}
        & \textbf{~~~~-0.8~~~~~~} & \textbf{~~~~~~+0.8~~~~} 
        & \textbf{~~~~-0.6~~~~~~} & \textbf{~~~~~~+0.6~~~~} \\
        \midrule
        StreamVLN 
        & 52.6{\scriptsize\,\textcolor{red!70!black}{($\downarrow$4.3)}} 
        & 53.7{\scriptsize\,\textcolor{red!70!black}{($\downarrow$3.2)}} 
        & 52.6{\scriptsize\,\textcolor{red!70!black}{($\downarrow$4.3)}} 
        & 49.5{\scriptsize\,\textcolor{red!70!black}{($\downarrow$7.4)}} \\
        
        JanusVLN  
        & 56.5{\scriptsize\,\textcolor{red!70!black}{($\downarrow$4.0)}} 
        & 57.0{\scriptsize\,\textcolor{red!70!black}{($\downarrow$3.5)}} 
        & 49.9{\scriptsize\,\textcolor{red!70!black}{($\downarrow$10.6)}} 
        & 50.5{\scriptsize\,\textcolor{red!70!black}{($\downarrow$10.0)}} \\
        
        StereoNav 
        & 69.0{\scriptsize\,\textcolor{red!70!black}{($\downarrow$3.8)}} 
        & 69.6{\scriptsize\,\textcolor{red!70!black}{($\downarrow$3.2)}} 
        & 69.4{\scriptsize\,\textcolor{red!70!black}{($\downarrow$3.4)}} 
        & 67.5{\scriptsize\,\textcolor{red!70!black}{($\downarrow$5.3)}} \\
        \bottomrule
    \end{tabular}
    }
\end{table}

As shown in Table~\ref{tab:robustness_perturbation_fluctuation}, lighting perturbations and height fluctuations cause smaller degradation than the severe disturbances reported in the main paper, but clear robustness differences remain across methods. Under lighting perturbations, StreamVLN drops by 4.3 and 3.2 points, while JanusVLN drops by 4.0 and 3.5 points. StereoNav exhibits comparable or smaller degradation, with drops of only 3.8 and 3.2 points, while maintaining higher absolute SR values of 69.0 and 69.6. Under camera-height fluctuations, the advantage of StereoNav becomes more evident. StreamVLN decreases by 4.3 and 7.4 points, and JanusVLN suffers larger drops of 10.6 and 10.0 points, whereas StereoNav decreases by only 3.4 and 5.3 points. Across all four mild settings, StereoNav consistently achieves the highest SR, ranging from 67.5 to 69.6, and shows the smallest or tied-smallest performance degradation. These results indicate that mild lighting and viewpoint-height changes have limited impact on StereoNav, further supporting the effectiveness of stereo-based semantic, structural, and geometric modeling for stable navigation under realistic observation variations.

\begin{figure*}[htbp]
    \centering
    \includegraphics[width=0.65\linewidth]{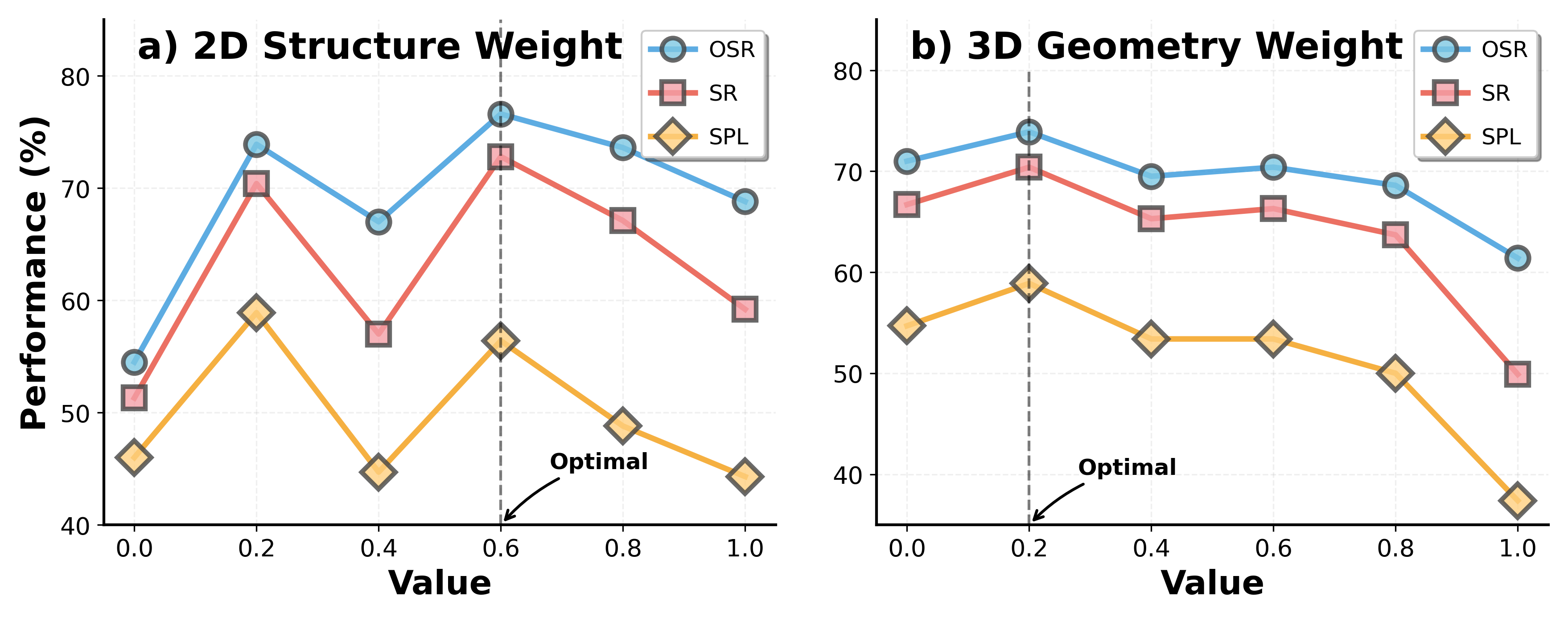}
    \caption{\textbf{Ablation study on fusion weights in Unified Understanding Modeling}. The results show that moderate structural guidance and lightweight geometric guidance lead to the best overall performance.}
    \label{fig:fusion_weight_ablation}
\end{figure*}

\section{Additional Ablation Studies}
\label{sec:Additional_Ablation_Studies}

We further study the fusion weights used in Unified Understanding Modeling. As shown in Fig.~\ref{fig:fusion_weight_ablation}, both the 2D structural branch and the 3D geometric branch are beneficial, but their contributions should be carefully balanced with the language-aligned semantic representation. For the 2D structural branch, a moderate weight of 0.6 yields the best overall performance, indicating that structural cues such as scene layout, object boundaries, and local spatial relations provide strong complementary information for navigation. However, further increasing this weight leads to performance degradation, suggesting that excessive structural emphasis may weaken semantic grounding. For the 3D geometric branch, the best performance is achieved with a smaller weight of 0.2. This shows that geometry is most effective as auxiliary depth-aware guidance, while over-weighting geometric tokens can introduce low-level matching noise and disturb high-level action prediction. Based on these observations, we adopt \(1.0\), \(0.6\), and \(0.2\) as the fusion weights for the semantic, structural, and geometric branches, respectively, which preserves semantic grounding as the dominant signal while incorporating sufficient structural and geometric cues for robust navigation.

% 预印文本时候注释掉
% \clearpage
% \newpage
% \input{checklist.tex}

\end{document}